\documentclass[10pt,twocolumn,letterpaper]{article}

\usepackage{3dv}
\usepackage{times}
\usepackage{epsfig}
\usepackage{graphicx}
\usepackage{amsmath}
\usepackage{amssymb}
\usepackage{subcaption}
\usepackage{tabularx}
\usepackage{booktabs} 
\usepackage{cuted}

\usepackage[pagebackref=true,breaklinks=true,letterpaper=true,colorlinks,bookmarks=false]{hyperref}

\threedvfinalcopy 

\def\threedvPaperID{22} 

\ifthreedvfinal\fi
\begin{document}

\title{MonoClothCap: Towards Temporally Coherent Clothing Capture \\ from Monocular RGB Video}

\author{
Donglai Xiang\textsuperscript{1}\thanks{Work partly done during internship at Facebook AI Research.} \qquad Fabian Prada\textsuperscript{2} \qquad Chenglei Wu\textsuperscript{2} \qquad Jessica Hodgins\textsuperscript{1,3}\\
\textsuperscript{1}Carnegie Mellon University \qquad \textsuperscript{2}Facebook Reality Labs Research \qquad \textsuperscript{3}Facebook AI Research\\
{\tt \small \{donglaix,jkh\}@cs.cmu.edu \qquad \{fabianprada,chenglei\}@fb.com}
}


\maketitle

\begin{abstract}
We present a method to capture temporally coherent dynamic clothing deformation from a monocular RGB video input. In contrast to the existing literature, our method does not require a pre-scanned personalized mesh template, and thus can be applied to in-the-wild videos. To constrain the output to a valid deformation space, we build statistical deformation models for three types of clothing: T-shirt, short pants and long pants. A differentiable renderer is utilized to align our captured shapes to the input frames by minimizing the difference in both silhouette, segmentation, and texture. We develop a UV texture growing method which expands the visible texture region of the clothing sequentially in order to minimize drift in deformation tracking. We also extract fine-grained wrinkle detail from the input videos by fitting the clothed surface to the normal maps estimated by a convolutional neural network. Our method produces temporally coherent reconstruction of body and clothing from monocular video. We demonstrate successful clothing capture results from a variety of challenging videos. Extensive quantitative experiments demonstrate the effectiveness of our method on metrics including body pose error and surface reconstruction error of the clothing.
\end{abstract}

\section{Introduction}
Dynamic capture of detailed human geometry and motion from monocular images and videos is attracting increasing attention in the computer vision and computer graphics community. High-quality human capture would enable applications in virtual and augmented reality, games, and movies. In recent years, great progress has been made on the estimation of general body shape from a single image or a monocular video \cite{kanazawa2018end,xu2019denserac,omran2018neural,pavlakos2019texturepose}. However, capturing the detailed deformation of clothing as it moves on the human body is still far from a solved problem.

Capturing a temporally coherent shape for clothing from monocular RGB imagery is an extremely challenging task, due to the fundamental ambiguity of single-view 3D reconstruction and the large deformation space of clothing. Previous work \cite{xu2018monoperfcap,habermann2019livecap,habermann2020deepcap} utilizes a 3D personalized actor model as a shape prior to track the dynamic clothing deformation. This model is acquired by multi-view reconstruction on an additional video of the same actor wearing the same clothing and rotating in a T-pose. However, such a model is generally unavailable for in-the-wild videos. The need for a pre-scanned template model limits the applicability of these approaches.

With the development of deep neural networks, other efforts have been made to regress a clothed human shape directly from a single input image with supervised learning \cite{varol2018bodynet,natsume2019siclope,saito2019pifu,tang2019neural,zheng2019deephuman,alldieck2019tex2shape,saito2020pifuhd}. These methods produce plausible results for individual input images of common human poses. However, it is difficult to extend them to capture temporally coherent dynamic clothing deformation from monocular videos for the following reasons. First, these methods are not robust to the variety of human motion due to the limited diversity of training data. They can easily produce incomplete geometry that is difficult to fix via post-processing. Second, it is non-trivial to estimate the temporal correspondence from the output of individual frames due to the data representation used (voxel \cite{varol2018bodynet,zheng2019deephuman}, depth map \cite{tang2019neural} or implicit function \cite{varol2018bodynet,natsume2019siclope,saito2019pifu,tang2019neural,zheng2019deephuman,alldieck2019tex2shape,saito2020pifuhd}). This limits the application of these methods in scenarios that require correspondence, such as clothing retargeting or image editing.

In this work, we present a novel method to capture dynamic clothing deformation from a monocular RGB video in a \textit{temporally coherent} manner, as illustrated by Fig.~\ref{fig:teaser}. To the best of our knowledge, it is the first attempt to solve this challenging problem without the prerequisite of a pre-scanned personalized template \cite{xu2018monoperfcap,habermann2019livecap,habermann2020deepcap}.

Our method is based on the following observations. First, a deformation model of the clothing that provides a statistical shape prior is key to solving the problem. It not only reduces the ambiguity of single-view 3D reconstruction, but also helps to estimate temporal correspondence across frames. While clothing models have been investigated in the existing literature \cite{yang2018analyzing,ma2020learning} for the purpose of clothing shape generation, our work is the first study that fully demonstrates the value of a clothing model for RGB-based clothing capture\footnote{Due to the limitation in types of available clothing data to train our model, in this paper, we assume that the subject to be captured wears a T-shirt on the upper body and shorts or pants on the lower body.}. Second, to solve the clothing capture problem, we make use of human appearance information including silhouette, segmentation, texture and surface normal. We present a novel method to integrate all those image measurements using a differentiable renderer \cite{liu2019soft,chen2019learning}. Our method captures the realistic dynamic of clothing in a temporally coherent manner including fine-grained wrinkle details from various videos.

\noindent \textbf{Our Contributions.} (1) We present the first approach for temporally coherent clothing capture from a monocular RGB video without using a pre-scanned template of the subject. (2) We propose a novel method to capture clothing deformation by fitting statistical clothing models to image measurements including silhouette, segmentation, texture and surface normal with a differentiable renderer.

\section{Related Work}

\noindent \textbf{Single-Image Human Pose and Shape Estimation.} Most previous work in human pose estimation focuses on the position of body keypoints in 2D \cite{wei2016convolutional,cao2017realtime,cao2019openpose} and 3D \cite{zhou2017towards,pavlakos2017coarse,sun2018integral}. Because estimating 3D pose from single images is highly ill-posed, deformable human models including SMPL \cite{loper2015smpl}, SMPL-X \cite{pavlakos2019expressive} and Adam \cite{joo2018total} are used to help with the problem by fitting the models to images \cite{bogo2016keep,pavlakos2019expressive,xiang2019monocular}. These models not only provide a strong prior for body pose, but also enable estimation of 3D body shape from single images. Deformable human models can be further integrated in deep neural network architectures \cite{kanazawa2018end,omran2018neural,pavlakos2019texturepose,xu2019denserac}. These networks are usually trained in a weakly-supervised manner without full 3D supervision.

Because deformable human models are not able to express clothing shape, all the work above only estimates body shapes with minimal clothing. Detailed clothing shape has been largely ignored in the previous literature, except a few papers \cite{varol2018bodynet,natsume2019siclope,saito2019pifu,zheng2019deephuman,tang2019neural,gabeur2019moulding}. These methods use deep neural networks to infer dense clothed human shapes in various data representation including voxels \cite{varol2018bodynet,zheng2019deephuman}, depth maps \cite{tang2019neural}, point clouds \cite{gabeur2019moulding} and implicit functions \cite{natsume2019siclope,saito2019pifu,saito2020pifuhd,huang2020arch}, all with supervised learning. However, because the amount of available training data is very limited, these methods are not robust to human motion. In addition, it is non-trivial to estimate correspondence across frames required for clothing capture due to their data representation. Our method achieves temporally coherent body and clothing capture in terms of both geometry and correspondence with the help of a statistical clothing model.

\noindent \textbf{Garment Modeling and Reconstruction.} Human clothing, especially physically based simulation of garments \cite{baraff1998large,bridson2003simulation,volino2009simple,sigal2015perceptual,li2018implicit}, has been extensively studied due to its important role in animation. Recently, there is growing interest in modeling garments in a data-driven manner. Pons-Moll \textit{et al.} \cite{pons2017clothcap} proposes a method to automatically segment 4D clothed human scans into different garment pieces, and track the deformation of each piece over time. The captured clothing data can be further used to train a deformable model, either a linear model \cite{yang2018analyzing} or a deep neural network \cite{ma2020learning}. In those methods, the clothing models are primarily used for shape generation, while we use the model to track clothing deformation from a monocular video.

Another line of work reconstructs clothing shape from images by allowing per-vertex deformation on top of the SMPL body model. Alldieck \textit{et al.}  \cite{alldieck2018video,alldieck2018detailed} builds clothed human avatar from videos of a person slowly rotating in A-pose. This is further improved to use only images of several different views \cite{alldieck2019learning,bhatnagar2019mgn} or even a single image \cite{alldieck2019tex2shape}. However, these methods reconstruct clothing as static objects without considering the temporal dynamics. By contrast, in this paper, we address the challenging problem of capturing clothing dynamics from a monocular video.

\begin{figure}[t!]
  \centering
  \includegraphics[width=\linewidth]{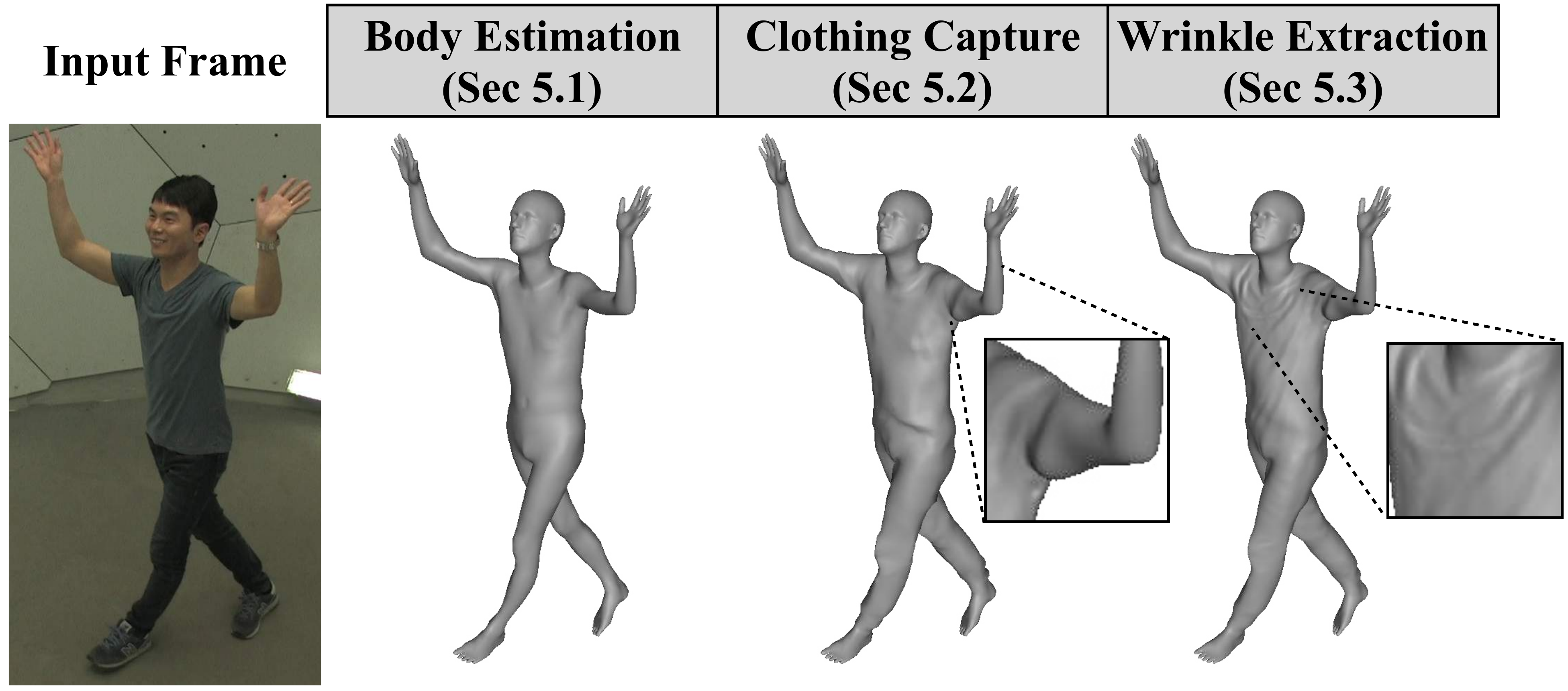}
  \caption{An overview of our clothing capture pipeline.}
  \label{fig:overview}
\end{figure}

\noindent \textbf{Monocular Human Performance Capture.} Motion capture and performance capture refer to the capture of space-time coherent human motion sequences in the form of sparse 3D joints and surface geometry respectively. Many approaches have been developed to enable motion and performance capture from multi-view inputs \cite{rhodin2016general,robertini2016model,robertini2018illumination,huang2018deep}. Here we focus on monocular-based capture methods. Mehta \textit{et al.} \cite{mehta2017vnect,mehta2020xnect} proposes systems to capture body skeleton motion from a single RGB video in real time. Some performance capture methods \cite{yu2018doublefusion,yu2019simulcap} are proposed to capture dense human body and clothing geometry from a monocular RGB-D video using a double-layer representation. Most relevant to our work are performance capture methods from monocular RGB videos \cite{xu2018monoperfcap,habermann2019livecap}. These methods, however, require a pre-scanned mesh template of the subject, which restricts the applications where they can be used. Habermann \textit{et al.} \cite{habermann2020deepcap} further proposes to train a deep neural network to deform a pre-scanned mesh template to match the surface deformation in the video. This method, requires the mesh template and multi-view images of the subject for network training. Our method relaxes the constraint to scenarios such as in-the-wild videos where neither pre-scanned templates nor multi-view images are available.

\section{Method Overview}

In this section, we present an overview of our approach. Our goal is to capture the dynamic deformation of three types of garments, T-shirt, shorts and pants, along with the underlying body shape from a monocular video. Our method takes as input a sequence of images, denoted as $\{\mathbf I_i\}_{i=1}^F$, where $F$ is the number of frames in the sequence. The subject is assumed to be wearing a T-shirt for the upper body. The clothing for the lower body is manually identified as either short pants or long pants. Our method outputs a sequence of mesh pairs $\{\mathbf M^b_i, \mathbf M^c_i\}_{i=1}^F$, where $\mathbf M^b_i$ denotes the body mesh and $\mathbf M^c_i$ denotes the clothed mesh. $\{\mathbf M^b_i\}_{i=1}^F$ and $\{\mathbf M^c_i\}_{i=1}^F$ are both temporally coherent with fixed topology across time. $\mathbf M^b_i$ and $\mathbf M^c_i$ share the same vertex positions except for the clothing region.

Our method makes use of linear clothing deformation models defined in the canonical pose. We briefly describe our model formulation and model building procedure in Section \ref{sec:model}. Our pipeline to capture clothing from a monocular video consists of four stages, explained in Section \ref{sec:pipeline}. First, we estimate the underlying body pose and shape of subject (Section \ref{sec:body}). Then, we run sequential tracking of the clothing using our linear clothing models. This step is followed by a batch optimization stage including all the frames to produce temporally coherent dynamic clothing deformation (Section \ref{sec:clothing}). In the final stage, we add fine-grained wrinkle detail to our results (Section \ref{sec:sfs}). A visualization of this pipeline is shown in Fig.~\ref{fig:overview}.

\begin{figure*}[t!]
  \centering
  \includegraphics[width=\linewidth]{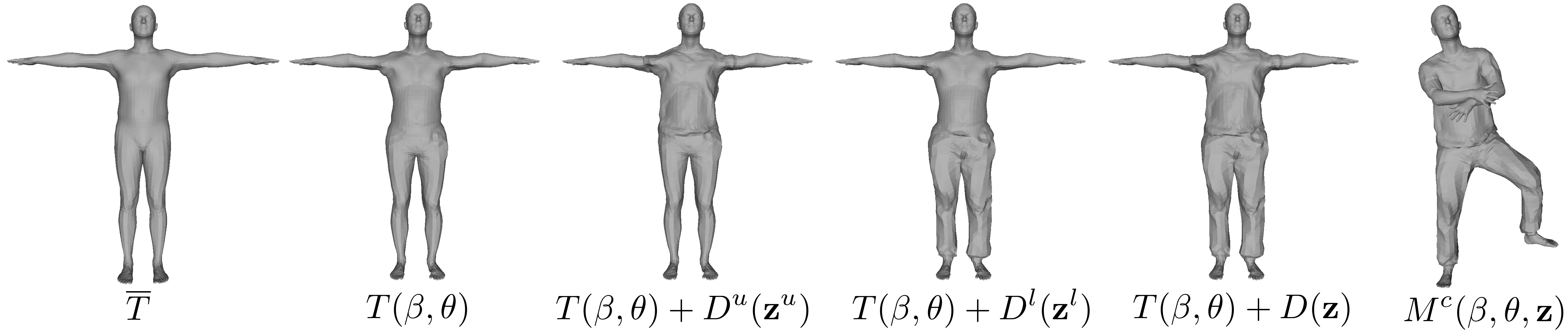}
  \caption{A visualization of our clothing model formulation. From left to right, we show (1) the mean SMPL template, (2) personalized body shape with pose dependent deformation, (3) upper clothing offsets, (4) lower clothing offsets, (5) combined clothing offsets, and (6) posed clothing output.}
  \label{fig:clothModels}
\end{figure*}

\section{Statistical Clothing Deformation Model}
\label{sec:model}

Statistical models of clothing have been investigated for clothing shape generation in the previous literature \cite{yang2018analyzing,ma2020learning}, but have yet to be exploited for capturing clothing from a monocular video. In this section we give the mathematical formulation of our clothing deformation models and briefly describe the procedure to learn these models from data.

\subsection{Model Formulation}

Our clothing models are built on top of the SMPL body model \cite{loper2015smpl}. SMPL is controlled by a set of model parameters $(\boldsymbol\beta, \boldsymbol\theta)$, where $\boldsymbol\beta \in \mathbb R^{10}$ is the shape coefficients and $\boldsymbol\theta \in \mathbb R^{72}$ is the joint angles that control body pose. We denote the set of $n_v=6890$ output vertices by $M(\boldsymbol\beta, \boldsymbol\theta)$. Then formally,
\begin{align}
    M(\boldsymbol\beta, \boldsymbol\theta) = W(T(\boldsymbol\beta, \boldsymbol\theta), J(\boldsymbol\beta), \boldsymbol\theta, \mathcal W),
\end{align}
where $W$ is the Linear Blend Skinning (LBS) function; $T(\boldsymbol\beta, \boldsymbol\theta)$ is the rest pose body shape; $J(\boldsymbol\beta)$ is the locations of $24$ kinematic joints; $\mathcal W$ is the blend weights. In particular, the unposed shape $T(\boldsymbol\beta, \boldsymbol\theta)$ is defined as the sum of template shape $\overline T$, shape dependent deformation $B^S(\boldsymbol{\beta})$ and pose dependent deformation $B^P(\boldsymbol\theta)$,
\begin{align}
    T(\boldsymbol\beta, \boldsymbol\theta) = \overline{T} + B^S(\boldsymbol\beta) + B^P(\boldsymbol\theta)
\end{align}

On top of the SMPL model, an extra additive offset field $D$ is introduced to account for clothing deformation in rest pose, i.e.,
\begin{align}T^c = T(\boldsymbol\beta, \boldsymbol\theta) + D.\end{align}
$D$ includes a number of $n_v$ per-vertex offsets, each denoted by $D_j \in \mathbb R^3$, where $1 \le j \le n_v$. Here we decompose $D$ into offsets from the upper clothing $D^u$ and offsets from the lower clothing $D^l$. $D^u$ and $D^l$ share the same dimensionality as $D$. They take non-zero values if the respective garments cover body vertex $j$; for an exposed skin vertex we have $D^u_j = D^l_j = 0$. Notice that some body vertices might be covered by both upper and lower clothing, for example, around the waist. To account for this phenomenon, we merge $D^u$ and $D^l$ into a single offset field $D$ by
\begin{gather}
    D_j =
    \begin{cases}
    D^u_j & \text{if } \Vert D^u_j \Vert \ge \Vert D^l_j \Vert, \\
    D^l_j & \text{otherwise.}
    \end{cases}
\end{gather}

The dimensions of $D^u$ and $D^l$ are very high ($3n_v$), so we use PCA dimension reduction to enable control with low-dimensional parameters $\mathbf z^u, \mathbf z^l \in \mathbb R^{n_z}$. Formally,
\begin{align} D^k(\mathbf z^k) = \mathbf A^k \mathbf z^k + \overline{\mathbf{d}^k}, ~ k \in \{u,l\} \end{align}
where $\mathbf A^k \in \mathbb R^{3n_v \times n_z}$ is the matrix of PCA bases and $\overline{\mathbf{d}^k}$ is the mean value vector. We use the skinning function $W$ of SMPL to transform the clothed shape from rest pose to target pose. Finally, our clothing model is formulated as
\begin{gather}
    M^c(\boldsymbol{\beta}, \boldsymbol{\theta}, \mathbf{z}) = W( T^c(\boldsymbol{\beta}, \boldsymbol{\theta}, \mathbf{z}), J(\boldsymbol\beta), \boldsymbol\theta, \mathcal W ), \\
    T^c(\boldsymbol{\beta}, \boldsymbol{\theta}, \mathbf{z}) = T(\boldsymbol\beta,\boldsymbol\theta) + D(\mathbf z),
\end{gather}
where $\mathbf z = \{\mathbf z^u, \mathbf z^l\}$ is the collection of clothing parameters. A visual illustration of our clothing model formulation is shown in Figure \ref{fig:clothModels}.

\subsection{Model Building}
We build our models from the BUFF dataset \cite{zhang2017detailed}, a collection of high-resolution 4D people scan. We build a model for each of the three garment types in the dataset, T-shirts, shorts and pants. For each garment type $k$, we need to train the model parameters $\{\mathbf A^k, \overline{\mathbf d^k}\}$ from a collection of clothing offsets, denoted by $\mathbf X^k \in \mathbb R^{3n_v \times n_k}$, where $n_k$ is the number of samples for garment type $k$ in the dataset. To obtain each sample in the collection, we follow \cite{zhang2017detailed,pons2017clothcap} to register the raw scan with SMPL model. This operation not only brings the raw scan data into the same topology, but also ``unposes'' the human shape with clothing into the rest pose, denoted by $\mathbf X^c$. We also follow \cite{zhang2017detailed} to estimate the underlying body shape of the subject in rest pose, denoted by $\mathbf X^b$. In addition, we obtain a per-vertex binary mask $\boldsymbol \sigma^k$ that has value $1$ for the region of garment type $k$ and $0$ for any other regions (skin and other clothing types) by rendering the meshes to images and applying a state-of-the-art clothing segmentation algorithm \cite{gong2019graphonomy}. Then we obtain the clothing offset data $\mathbf X^k$ for garment type $k$ by
\begin{align}
    \mathbf X^k = (\mathbf X^c - \mathbf X^b) \odot \boldsymbol \sigma^k,    
\end{align}
where $\odot$ denotes the element-wise multiplication. We use a standard PCA training algorithm based on Singular Value Decomposition (SVD), leaving $n_z = 50$ bases in our model. We refer readers to the original papers \cite{zhang2017detailed,pons2017clothcap} for details on scan registration and underlying body shape estimation.

\section{Monocular Clothing Capture}
\label{sec:pipeline}

Given the pre-trained clothing models, we now present our approach for temporally coherent clothing capture from only a monocular video.

\subsection{Body Motion Estimation}
\label{sec:body}

In the first stage, we estimate underlying body motion in 3D with the SMPL body model. We estimate per-frame SMPL pose parameters $\boldsymbol \theta_i$ and global translation $\mathbf t_i \in \mathbb R^3$, together with SMPL shape parameters $\boldsymbol \beta$ across the whole sequence. Meanwhile, we estimate the camera intrinsics $\mathbf K$ of a full perspective projection model for all the frames. In order to achieve good robustness under different in-the-wild scenarios, we integrate a variety of different image measurements into an energy optimization problem. Formally, we solve the following minimization problem:
\begin{align}
    \min_{ \boldsymbol\beta, \{\boldsymbol\theta_i, \mathbf t_i\}_{i=1}^F, \mathbf K } E^b = E^b_{\text{2d}} + E^b_{\text{dp}} + E^b_{\text{sil}} + E^b_{\text{pof}} + E^b_\text{reg}.
    \label{eq:body-energy}
\end{align}
In particular, $E^b_{\text{2d}}$ is the squared $L_2$ error between projected SMPL joints and 2D keypoint detection from OpenPose \cite{wei2016convolutional,cao2017realtime,cao2019openpose}. $E^b_{\text{dp}}$, also used in Guler \textit{et al.} \cite{guler2019holopose}, is an energy term for dense correspondence estimation from DensePose \cite{alp2018densepose}. Specifically, for any pixel $\mathbf p$ in the image with DensePose prediction, we identify the corresponding SMPL vertex index $j(\mathbf p)$ and optimize an energy term defined as
\begin{align}
    E^b_{\text{dp}} = \frac1F \sum_{i=1}^F \sum_{\mathbf p} \Vert \Pi(M_{j(\mathbf p)}(\boldsymbol \beta, \boldsymbol \theta_i) + \mathbf t_i; \mathbf K) - \mathbf p \Vert^2,
\end{align}
where $\Pi$ denotes the projection function determined by the camera intrinsics $\mathbf K$. $E^b_{\text{sil}}$ is the silhouette matching term. We extract silhouettes $\mathbf S_i$ of our SMPL body mesh with a differentiable renderer \cite{liu2019soft}, and obtain the target silhouette $\hat{\mathbf S}_i$ from a clothing segmentation algorithm \cite{gong2019graphonomy}. We use an Intersection-over-Union error \cite{liu2019soft}
\begin{align}
    E^b_{\text{sil}} = \frac1F \sum_{i=1}^F \left( 1 -\frac{\Vert \mathbf S_i \odot \hat{\mathbf S}_i \Vert_1}{\Vert \mathbf S_i + \hat{\mathbf S}_i - \mathbf S_i \odot \hat{\mathbf S}_i \Vert_1} \right).
\end{align}
$E^b_{\text{pof}}$ is an error term based on 3D orientation between adjacent joints in the body skeleton hierarchy. We match the spatial orientation of SMPL body joints to the prediction of Part Orientation Field (POF) similar to \cite{xiang2019monocular}. We refer readers to the original papers \cite{xiang2019monocular,luo2018orinet} for details. We also apply regularization on our estimation, denoted by $E^b_{\text{reg}}$, which consists of a Mixture of Gaussian prior for body pose $\{\boldsymbol{\theta}_i\}_{i=1}^F$ \cite{bogo2016keep}, $L_2$ regularization on the shape parameters $\boldsymbol \beta$, and temporal smoothness terms to reduce motion jitters. 

After solving the energy optimization, we obtain a temporally consistent body mesh for every frame by $\mathbf M^b_i = M(\boldsymbol\beta, \boldsymbol\theta_i)$. We fix the SMPL parameters $\boldsymbol\beta, \{\boldsymbol\theta_i,\mathbf t_i\}_{i=1}^F$ and camera parameters $\mathbf K$ during later stages of our pipeline. The estimated body meshes provide a strong guidance for the subsequent estimation of clothing deformation.

\subsection{Clothing Deformation Capture}
\label{sec:clothing}

We now illustrate our proposed method to capture clothing deformation. Compared to previous work \cite{xu2018monoperfcap,habermann2019livecap,habermann2020deepcap} where a pre-scanned template of the subject is assumed, this problem is significantly more challenging due to the lack of strong shape prior to resolve the single-view 3D ambiguity, and the lack of a pre-defined personalized texture that provides correspondence for surface tracking. To solve this problem, we (1) exploit the deformation space learned in our clothing models and (2) progressively extract a personalized texture from the input image sequence to enable surface tracking across time and reduce drifting.

\begin{figure}[t]
  \centering
  \includegraphics[width=\linewidth]{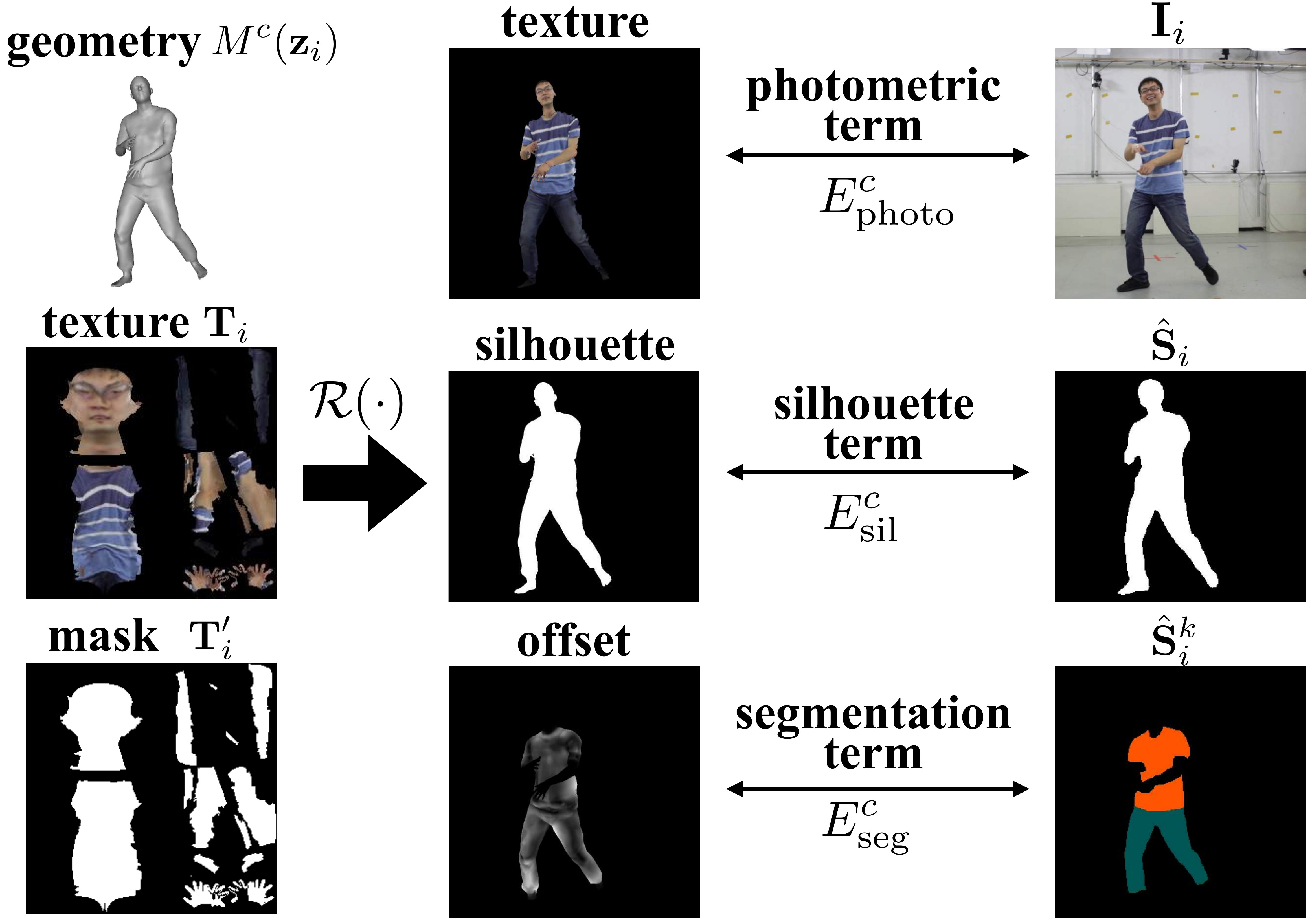}
  \caption{Explanation of the energy terms used for clothing capture. We obtain the rendered texture, silhouette and clothing offset with a differentiable renderer, which are compared with target images using different energy terms.}
  \label{fig:energy}
\end{figure}

We perform clothing capture in a sequential manner. For each frame $i$, we estimate per-frame clothing parameters for clothing on the upper and lower body $\mathbf z_i = \{\mathbf z^u_i, \mathbf z^l_i\}$ given the input image $\mathbf I_i$, initializing from the result of the previous frame $\mathbf z_{i-1}$. We formulate the task as solving an energy optimization problem, formally,
\begin{align}
    \min_{\mathbf z_i} E^c = E^c_{\text{sil}} + E^c_{\text{seg}} + E^c_{\text{photo}} +  E^c_{\text{reg}}.
    \label{eq:clothing-sequential}
\end{align}
Now we explain each cost term individually. An illustration of the different cost terms is shown in Fig.~\ref{fig:energy}.

\noindent\textbf{Silhouette matching term} $E^c_{\text{sil}}$: Similar to $E^b_{\text{sil}}$ in the first stage, we use a differentiable renderer to match the silhouette of our rendering output with a target silhouette extracted from the original image. Differently, here what we compare with target silhouette is the silhoutte of human shape with clothing $M^c(\boldsymbol\beta, \boldsymbol\theta_i, \mathbf z_i)$, instead of bare body shape $M(\boldsymbol\beta, \boldsymbol\theta_i)$ in the previous stage.

\noindent\textbf{Clothing segmentation term} $E^c_{\text{seg}}$: Clothing segmentation \cite{gong2019graphonomy} provides not only the overall silhouette of the person, but also the boundary between different garment regions and exposed skin in the image. We utilize this information by penalizing the clothing offset on vertices whose projection falls outside the segmentation region. Concretely, the differentiable renderer is used to render the per-vertex offset fields $D^u(\mathbf z^u_i), D^l(\mathbf z^l_i)$ on the clothed mesh $M^c(\mathbf z_i)$\footnote[2]{SMPL body parameters $\boldsymbol\beta, \boldsymbol\theta_i$ and $\mathbf t_i$ are fixed in this stage thus omitted here.}; we denote the output by $\mathcal R\left(D^k(\mathbf z^k_i) \right)$, where $k$ represents either $u$ for upper clothing or $l$ for lower clothing. In addition, we denote the segmentation masks for the clothing region $k \in \{u,l\}$ by $\hat{\mathbf S}^k_i$. Then we have

\begin{align}
    E^c_{\text{seg}} = \sum_{k \in \{u,l\}} \sum_{\mathbf p} (1 - \hat{\mathbf S}^k_i) \odot \left\Vert \mathcal R\left(D^k(\mathbf z^k_i) \right) \right\Vert^2,
\end{align}
where $\mathbf p$ iterates over all pixels in the image. Effectively, for each clothing type we penalize the clothing offset outside the corresponding clothing region, where $\hat{\mathbf S}^k_i$ is $0$. The gradient in the image domain is propagated to the mesh by the differentiable renderer $\mathcal R$.

\noindent\textbf{Photometric tracking term} $E^c_{\text{photo}}$: This term is introduced to estimate temporal correspondence more accurately, especially when the garments we capture have high-contrast texture. We progressively build a personalized RGB texture image $\mathbf T_{i}$ in a pre-defined UV space of SMPL model, along with a binary mask $\mathbf T'_{i}$ that indicates texels where RGB values in $\mathbf T_{i}$ have been identified. The photometric tracking term is defined to compare the rendered output of our clothing models using $\mathbf T_{i}$ with the input image. For this purpose we use a differential renderer that works with UV texture \cite{chen2019learning,jatavallabhula2019kaolin}, and denotes the rendered output as $\mathcal R\left(\mathbf T_{i} \right)$. We also render the mesh with $\mathbf T'_{i}$ to indicate pixels where texture from $\mathbf T_{i}$ is available. Formally, we have
\begin{align}
    E^c_{\text{photo}} = \sum_{\mathbf p} \Vert \mathcal R\left(\mathbf T_{i} \right) - \mathbf I_i \Vert^2 \odot \mathcal R\left(\mathbf T'_{i} \right),
\end{align}
where the summation is taken over the pixels in the image. After the optimization in Eq. \ref{eq:clothing-sequential} is solved for frame $i$, we update $\mathbf T_i, \mathbf T'_i$ to obtain $\mathbf T_{i+1}, \mathbf T'_{i+1}$, which will be used for solving optimization for frame $i+1$. To achieve this, we project $\mathbf I_i$ to the mesh surface and fill in new RGB values to UV texels in $\mathbf T_i$ where no previous values have been identified, indicated by $0$s in $\mathbf T'_i$. Corresponding texels in $\mathbf T'_i$ are also set to $1$ to obtain $\mathbf T'_{i+1}$. This process is initialized by setting $\mathbf T_1$ and $\mathbf T'_1$ to $0$; in other words, the photometric tracking term takes no effect for the first frame in the sequence, since no texture has been extracted.

\noindent\textbf{Regularization term} $E^c_{\text{reg}}$: Our clothing deformation models are PCA-based linear models. They may produce unreasonable shapes when the parameters $\mathbf z$ are large. Therefore we apply regularization on the cloth parameters using an adaptive cost function $\rho$ that penalizes large input values:
\begin{align}
    E^c_{\text{reg}} = \rho\left(\Vert \mathbf z_i \Vert^2\right).
\end{align}

The sequential tracking stage is then followed by a batch optimization stage that optimize for all $F$ frames in the sequence together. The energy function we use is the same as the previous stage (Eq. \ref{eq:clothing-sequential}) with an additional term that penalizes too drastic temporal change of clothing parameters, which helps to produce temporally stable results. This term is defined as
\begin{align}
    E^c_{\text{temp}} = \frac1{F-1} \sum_{i=1}^{F-1} \Vert \mathbf z_{i+1} - \mathbf z_{i} \Vert^2.
\end{align}
The output of batch optimization stage is a sequence of capture results with clothing $\{\mathbf M^c_i = M^c(\boldsymbol\beta, \boldsymbol\theta_i, \mathbf z_i)\}_{i=1}^F$.

\subsection{Wrinkle Detail Extraction}
\label{sec:sfs}

Up to the batch optimization stage, we can capture large clothing deformation. However, the results are limited by low mesh resolution and unable to capture the fine-grained wrinkles on the clothing. Therefore, the last stage of our approach is to extract wrinkle details from the original images and apply them to our coarsely tracked meshes.

Traditionally, such wrinkle details are captured with Shape from Shading (SfS) \cite{wu2013set,alldieck2018detailed}. For in-the-wild monocular clothing capture, we empirically find it difficult to extract wrinkles reliably by SfS due to complex garment albedo, large variation of lighting conditions and self-shadowing. Recently, we observed the success of learning-based approaches in estimating accurate surface normal for human appearance using neural networks \cite{tang2019neural,saito2020pifuhd}. The estimated surface normal provides strong and direct clues on how the wrinkles should be added to our clothing capture results in order to match the original images.

Formally, let us denote the output of a surface normal estimation network for frame $i$ to be $\mathbf I^n_i$, a 3-channel normal map for each pixel in the original image.
We first subdivide the mesh $\mathbf M^c_i$ with Loop subdivision to increase the spatial resolution, with the subdivided mesh denoted by $\mathbf M^s_i$. Then, we solve for a deformed mesh $\mathbf O_i$ whose rendered normal map matches the estimated normal map $\mathbf I^n_i$ in the garment region. We denote the rendered normal output by $\mathcal R^n(\mathbf O_i)$, where $\mathcal R^n$ is the differential renderer. We solve the following optimization problem for each frame $i$ individually:

\begin{gather}
    \min_{\mathbf O_i} E^w = E^w_{\text{normal}} + E^w_{\text{reg}} + E^w_{\text{lpl}}, \\
    E^w_{\text{normal}} = \sum_{\mathbf p \in \hat{\mathbf S}^c_i} \Vert \nabla\mathcal R^n(\mathbf O_i) - \nabla\mathbf I^n_i \Vert^2, \\
    E^w_{\text{reg}} = \Vert \mathbf O_i - \mathbf M^s_i \Vert^2, ~
    E^w_{\text{lpl}} = \left\Vert \mathbf L \mathbf O_i \right\Vert^2.
\end{gather}
where $\nabla$ denotes the image gradient operator, and $\mathbf L$ denotes the mesh Laplacian operator, $\hat {\mathbf S}^c_i = \hat{\mathbf S}^u_i \bigcup \hat{\mathbf S}^l_i$ is the union of all pixels in the clothing segmentation masks. Here we penalize the difference between normal maps in the image gradient domain to be more tolerant to error in absolute normal direction from the neural network. We also restrict the deformation of $\mathbf O_i$ from $\mathbf M^s_i$ be in the direction of the camera rays. Here, we use the implementation of differentiable renderer in \cite{chen2019learning,jatavallabhula2019kaolin} and surface normal network in \cite{saito2020pifuhd}. The final results are the deformed meshes $\{\mathbf O_i\}_{i=1}^F$.

\section{Quantitative Evaluation}

\begin{figure}[t]
  \centering
  \includegraphics[width=\linewidth]{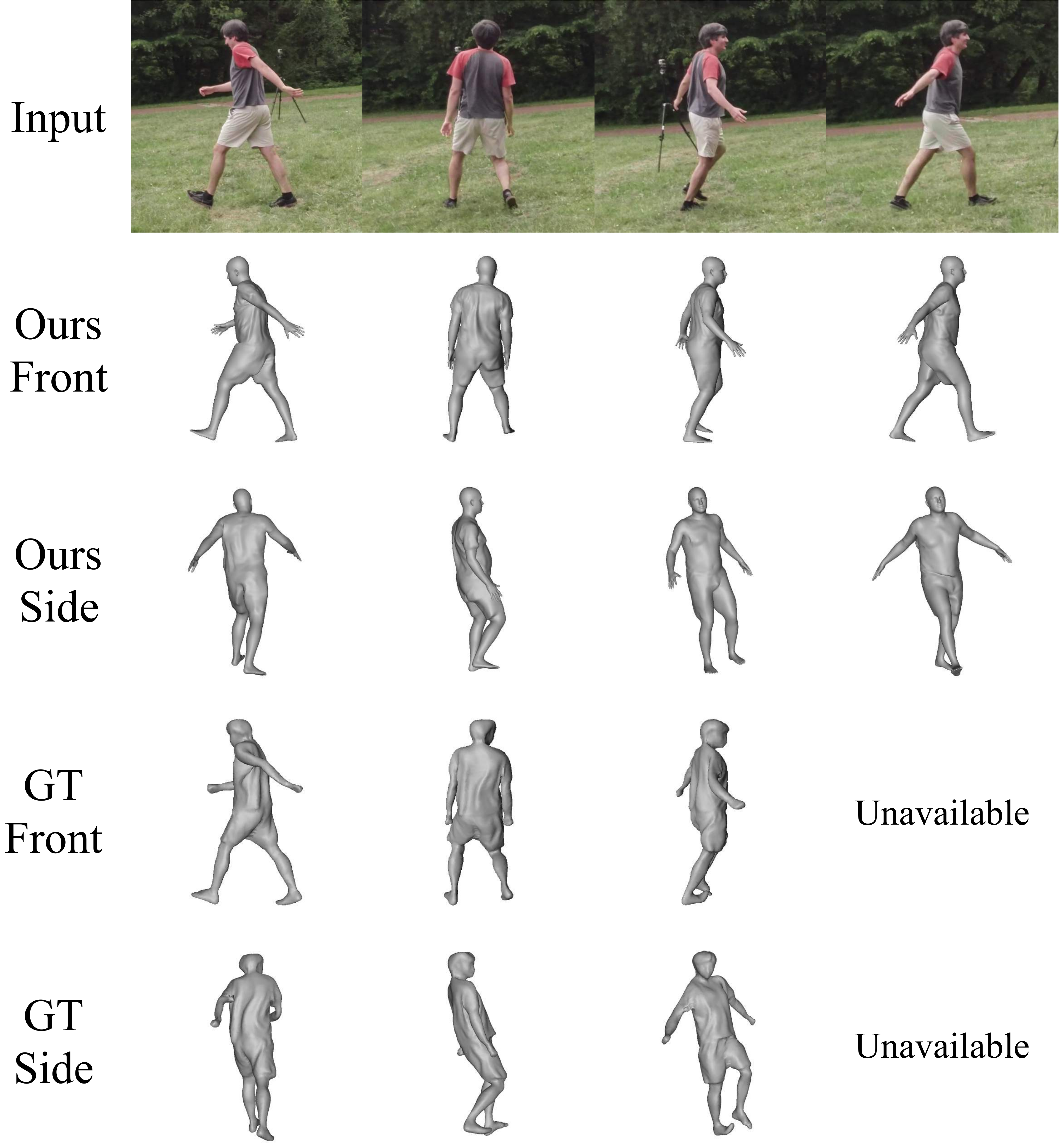}
  \caption{Visualization of experiment results on the \textit{Pablo} sequence. From top to bottom we show original images, our results from the front and side views, and ground truth from the front and side views. The ground truth mesh for the last frame is not provided in the dataset.}
  \label{fig:pablo-qualitative}
\end{figure}


In this section, we present the results of quantitative evaluation. We use a benchmark sequence from the MonoPerfCap dataset \cite{xu2018monoperfcap} and video sequences rendered from BUFF dataset \cite{zhang2017detailed} to test the performance of our method.

\subsection{Evaluation on MonoPerfCap Dataset}
\label{sec:eval-monoperfcap}

\noindent\textbf{Experiment Setting.} We follow previous work \cite{xu2018monoperfcap} to use the \textit{Pablo} sequence in their dataset to perform quantitative comparison. Surface meshes and 3D joints obtained by a multi-view performance capture method \cite{robertini2016model} are provided as the ground truth in the dataset. We compare our method with a state-of-the-art template-based performance capture method \cite{xu2018monoperfcap}\footnote[3]{Monocular capture results are provided by the author.} and single-image human reconstruction methods \cite{zheng2019deephuman,saito2019pifu,saito2020pifuhd,alldieck2019tex2shape,zhu2019detailed}. Body pose is not estimated in \cite{alldieck2019tex2shape}, so we apply our estimated body pose to their T-pose results.

\noindent\textbf{Evaluation of Clothing Surface Reconstruction.} We first evaluate our method using a surface reconstruction metric. Due to the intrinsic depth-scale ambiguity of single-view reconstruction, we compute a global scaling factor from our result to the ground truth, which is applied to our result before comparison. Following \cite{xu2018monoperfcap} we align our results to the ground truth with a translation to eliminate the global depth offsets. We compute the average point-to-surface distance from all the ground truth vertices in the clothing region to the output mesh as the evaluation metric. The clothing region (the T-shirt and shorts) is obtained by manual segmentation on the ground truth surface mesh. The same procedure is applied to all the methods under evaluation. A visualization of our results is shown in Fig.~\ref{fig:pablo-qualitative}.

We report the mean surface error averaged across all frames in the middle column of Table \ref{table:pablo-error}. For the visualization of per-frame error curves please refer to our supplementary material. Our method achieves a significantly lower surface error compared to all previous single-image surface reconstruction methods. Our performance even comes close to the template-based tracking method \cite{xu2018monoperfcap} which requires a pre-scanned personalized template that provides strong prior information about the body and clothing shape of the subject. By contrast, our method does not require a pre-processed template, and therefore can be applied to a wider range of videos.

\begin{table}[t]
\centering
\begin{tabular}{p{3cm} c c}
    \hline\addlinespace[1pt]
    Methods & Surface Error & Joint Error  \\
    \hline\addlinespace[1pt]
    MonoPerfCap* \cite{xu2018monoperfcap} & \textbf{14.6} & 118.7 \\
    \hline\addlinespace[1pt]
    HMD \cite{zhu2019detailed} & 31.9 & - \\
    Tex2Shape \cite{alldieck2019tex2shape} & 27.7 & - \\
    DeepHuman \cite{zheng2019deephuman} & 24.2 & - \\
    PIFu \cite{saito2019pifu} & 30.5 & - \\
    PIFuHD \cite{saito2020pifuhd} & 26.5 & - \\
    Ours & \textbf{17.9} & \textbf{77.3} \\
    \hline
\end{tabular}
\caption{Quantitative comparison with previous work on \textit{Pablo} sequence using mean point-to-surface error and mean joint error across frames. All numbers are in mm. The method annotated with `*' uses a pre-scanned personalized template that provides a strong shape prior. Please see our supplementary material for per-frame error.}
\label{table:pablo-error}
\end{table}

\noindent\textbf{Evaluation of 3D Pose Estimation.} 
Although body pose estimation is not a focus of this paper, we follow \cite{xu2018monoperfcap} to validate our method on the metric of 3D joint error on the \textit{Pablo} sequence. Average per-joint 3D position error after alignment with translation is reported in Table \ref{table:pablo-error} (right). Our method achieves an error of $77.3$ mm, significantly lower than $118.7$ mm in \cite{xu2018monoperfcap}. This verifies the effectiveness of our body pose initialization that utilizes various image measurements including 2D joints, dense correspondences, silhouette, etc.


\subsection{Evaluation on BUFF Dataset}

\begin{figure}[t]
  \centering
  \includegraphics[width=\linewidth]{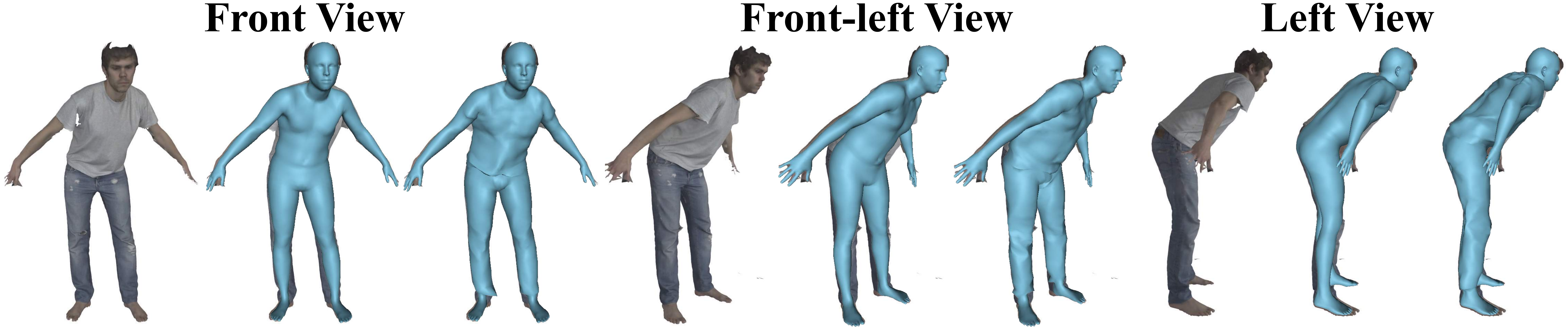}
  \caption{Visualization of three viewpoints in the BUFF evaluation. In each view we show the input image, body estimation result, and clothing capture result.}
  \label{fig:eval-buff-qualitative}
\end{figure}

\textbf{Experiment Setting.} BUFF \cite{zhang2017detailed} is a dataset of high-resolution 4D textured scan sequences of five people. In this experiment, we sample a test sequence from the BUFF dataset (00096-shortlong\_hips, first 200 frames) and train a pair of upper and lower clothing models with the data of four other people. We render the sequence from three views: front, left and front-left, as visualized in Fig.~\ref{fig:eval-buff-qualitative}. We evaluate our method in four stages: body initialization, sequential tracking, batch optimization and wrinkle extraction
The evaluation protocol is the same as Section \ref{sec:eval-monoperfcap}: we rigidly align the estimated and ground truth meshes with a global scaling and translation, and compute average distance from ground truth clothing vertices to our results.

\begin{figure*}[t]
  \centering
  \includegraphics[width=1.0\linewidth]{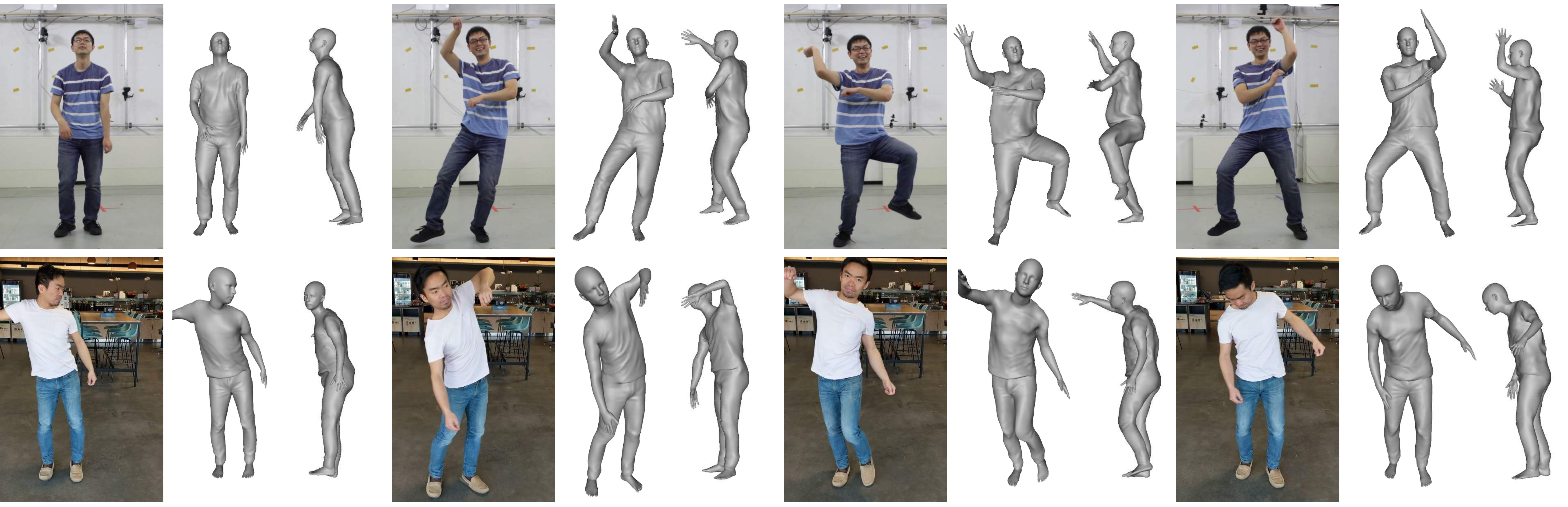}
  \caption{Examples of our clothing capture results. In each example, we show the input image, capture results from the front view and the side view. Please see our supplementary video for complete results.}
  \label{fig:qualitative}
\end{figure*}

\begin{table}[t]
\centering
\begin{tabular}{c c c c}
    \hline\addlinespace[1pt]
     & Front & Front-left & Left  \\
    \hline\addlinespace[1pt]
    Body only & 29.4 & 30.3 & 29.2 \\
    Clothed w/o batch & 26.8 & 25.5 & \textbf{24.7} \\
    Clothed w/ batch & \textbf{26.7} & \textbf{25.3} & \textbf{24.7} \\
    Clothed w/ wrinkle & 26.8 & 25.5 & 24.9 \\
    \hline
\end{tabular}
\caption{Quantitative ablation study of different stages of our method on the rendered BUFF dataset using mean point-to-surface error. All numbers are in mm. Please see our supplementary material for per-frame error.}
\label{table:eval-buff}
\end{table}


\noindent\textbf{Results.} The quantitative results are shown in Table \ref{table:eval-buff}. First, from all three viewpoints, results with clothing consistently achieve lower reconstruction error than body only. This verifies that our method captures clothing shape that cannot be explained by the SMPL body shape space. Second, we can see that temporal smoothing and wrinkle extraction, which improve the visual quality as shown in qualitative results, have little influence on the reconstruction error. Third, our results show similar range of error in the clothing region across different views, implying that our method is not very sensitive to the viewpoint variation.

\section{Qualitative Evaluation}

We qualitatively evaluate our method on various videos including public benchmark and in-the-wild videos where no pre-scanned template is available. Example results are shown in Figure \ref{fig:qualitative}. Please see our supplementary video for full results and qualitative comparison with other work.

As shown in the supplementary video, our result not only demonstrates better temporal robustness than the single-image 3D human reconstruction methods in terms of reconstructed surfaces, but also provides 3D temporal correspondences effectively shown by the re-rendering of our output mesh with a consistent texture map. This is hard to obtain by methods that regress 3D shape in voxels \cite{varol2018bodynet,zheng2019deephuman}, depth maps \cite{tang2019neural} or implicit functions \cite{natsume2019siclope,saito2019pifu,saito2020pifuhd}. Template-based monocular performance capture methods \cite{xu2018monoperfcap,habermann2019livecap} rely heavily on non-rigid surface regularization such as As-Rigid-As-Possible (ARAP), which often prevents those methods from capturing natural dynamics of the clothing deformation. In comparison, our method is able to capture more realistic dynamics of the garment with regularization provided by the clothing models.

In addition, we perform extensive ablation studies on various loss terms used in our pipeline. Please refer to our supplementary document for the results.

\section{Conclusion and Future Work}

In this paper, we have presented a method to capture temporally coherent dynamic deformation of clothing from a monocular video. To the best of our knowledge, we have shown the first result of temporally coherent clothing capture from a monocular RGB video without using a pre-scanned template. Our results on various in-the-wild videos endorse the effectiveness and robustness of our method.

Our method is limited by the types of garments in the available training data. We have demonstrated results on several types of tight clothing. Treatment of free-flowing garments like skirts requires collection of more data and additional design of the clothing model. Our method is constrained in the ability to capture drastically changing deformations due to the limited expressiveness of our models, which may be addressed by using higher-capacity models like a deep neural network. We also would like to further incorporate physics into the clothing models to enable more physically realistic clothing capture.


\noindent \textbf{Acknowledgements.} We would like to thank Eric Yu for his help with the rendering of our results using Blender.

{\small
\bibliographystyle{ieee}
\bibliography{egpaper_final}

\begin{thebibliography}{10}\itemsep=-1pt

\bibitem{alldieck2019learning}
T.~Alldieck, M.~Magnor, B.~L. Bhatnagar, C.~Theobalt, and G.~Pons-Moll.
\newblock Learning to reconstruct people in clothing from a single rgb camera.
\newblock In {\em Proceedings of the IEEE Conference on Computer Vision and
  Pattern Recognition}, pages 1175--1186, 2019.

\bibitem{alldieck2018detailed}
T.~Alldieck, M.~Magnor, W.~Xu, C.~Theobalt, and G.~Pons-Moll.
\newblock Detailed human avatars from monocular video.
\newblock In {\em 2018 International Conference on 3D Vision (3DV)}, pages
  98--109. IEEE, 2018.

\bibitem{alldieck2018video}
T.~Alldieck, M.~Magnor, W.~Xu, C.~Theobalt, and G.~Pons-Moll.
\newblock Video based reconstruction of 3d people models.
\newblock In {\em Proceedings of the IEEE Conference on Computer Vision and
  Pattern Recognition}, pages 8387--8397, 2018.

\bibitem{alldieck2019tex2shape}
T.~Alldieck, G.~Pons-Moll, C.~Theobalt, and M.~Magnor.
\newblock Tex2shape: Detailed full human body geometry from a single image.
\newblock In {\em Proceedings of the IEEE International Conference on Computer
  Vision}, pages 2293--2303, 2019.

\bibitem{alp2018densepose}
R.~Alp~G{\"u}ler, N.~Neverova, and I.~Kokkinos.
\newblock Densepose: Dense human pose estimation in the wild.
\newblock In {\em Proceedings of the IEEE Conference on Computer Vision and
  Pattern Recognition}, pages 7297--7306, 2018.

\bibitem{baraff1998large}
D.~Baraff and A.~Witkin.
\newblock Large steps in cloth simulation.
\newblock In {\em Proceedings of the 25th annual conference on Computer
  graphics and interactive techniques}, pages 43--54, 1998.

\bibitem{bhatnagar2019mgn}
B.~L. Bhatnagar, G.~Tiwari, C.~Theobalt, and G.~Pons-Moll.
\newblock Multi-garment net: Learning to dress 3d people from images.
\newblock In {\em {IEEE} International Conference on Computer Vision ({ICCV})}.
  {IEEE}, oct 2019.

\bibitem{bogo2016keep}
F.~Bogo, A.~Kanazawa, C.~Lassner, P.~Gehler, J.~Romero, and M.~J. Black.
\newblock Keep it smpl: Automatic estimation of 3d human pose and shape from a
  single image.
\newblock In {\em European Conference on Computer Vision}, pages 561--578.
  Springer, 2016.

\bibitem{bridson2003simulation}
R.~Bridson, S.~Marino, and R.~Fedkiw.
\newblock Simulation of clothing with folds and wrinkles.
\newblock In {\em Proceedings of the 2003 ACM SIGGRAPH/Eurographics symposium
  on Computer animation}, pages 28--36, 2003.

\bibitem{cao2019openpose}
Z.~Cao, G.~H. Martinez, T.~Simon, S.-E. Wei, and Y.~A. Sheikh.
\newblock Openpose: Realtime multi-person 2d pose estimation using part
  affinity fields.
\newblock {\em IEEE Transactions on Pattern Analysis and Machine Intelligence},
  2019.

\bibitem{cao2017realtime}
Z.~Cao, T.~Simon, S.-E. Wei, and Y.~Sheikh.
\newblock Realtime multi-person 2d pose estimation using part affinity fields.
\newblock In {\em Proceedings of the IEEE Conference on Computer Vision and
  Pattern Recognition}, pages 7291--7299, 2017.

\bibitem{chen2019learning}
W.~Chen, H.~Ling, J.~Gao, E.~Smith, J.~Lehtinen, A.~Jacobson, and S.~Fidler.
\newblock Learning to predict 3d objects with an interpolation-based
  differentiable renderer.
\newblock In {\em Advances in Neural Information Processing Systems}, pages
  9605--9616, 2019.

\bibitem{gabeur2019moulding}
V.~Gabeur, J.-S. Franco, X.~Martin, C.~Schmid, and G.~Rogez.
\newblock Moulding humans: Non-parametric 3d human shape estimation from single
  images.
\newblock In {\em Proceedings of the IEEE International Conference on Computer
  Vision}, pages 2232--2241, 2019.

\bibitem{gong2019graphonomy}
K.~Gong, Y.~Gao, X.~Liang, X.~Shen, M.~Wang, and L.~Lin.
\newblock Graphonomy: Universal human parsing via graph transfer learning.
\newblock In {\em Proceedings of the IEEE Conference on Computer Vision and
  Pattern Recognition}, pages 7450--7459, 2019.

\bibitem{guler2019holopose}
R.~A. Guler and I.~Kokkinos.
\newblock Holopose: Holistic 3d human reconstruction in-the-wild.
\newblock In {\em Proceedings of the IEEE Conference on Computer Vision and
  Pattern Recognition}, pages 10884--10894, 2019.

\bibitem{habermann2019livecap}
M.~Habermann, W.~Xu, M.~Zollhoefer, G.~Pons-Moll, and C.~Theobalt.
\newblock Livecap: Real-time human performance capture from monocular video.
\newblock {\em ACM Transactions on Graphics (TOG)}, 38(2):1--17, 2019.

\bibitem{habermann2020deepcap}
M.~Habermann, W.~Xu, M.~Zollhofer, G.~Pons-Moll, and C.~Theobalt.
\newblock Deepcap: Monocular human performance capture using weak supervision.
\newblock In {\em Proceedings of the IEEE/CVF Conference on Computer Vision and
  Pattern Recognition}, pages 5052--5063, 2020.

\bibitem{huang2018deep}
Z.~Huang, T.~Li, W.~Chen, Y.~Zhao, J.~Xing, C.~LeGendre, L.~Luo, C.~Ma, and
  H.~Li.
\newblock Deep volumetric video from very sparse multi-view performance
  capture.
\newblock In {\em Proceedings of the European Conference on Computer Vision
  (ECCV)}, pages 336--354, 2018.

\bibitem{huang2020arch}
Z.~Huang, Y.~Xu, C.~Lassner, H.~Li, and T.~Tung.
\newblock Arch: Animatable reconstruction of clothed humans.
\newblock In {\em Proceedings of the IEEE/CVF Conference on Computer Vision and
  Pattern Recognition}, pages 3093--3102, 2020.

\bibitem{jatavallabhula2019kaolin}
K.~M. Jatavallabhula, E.~Smith, J.-F. Lafleche, C.~F. Tsang, A.~Rozantsev,
  W.~Chen, and T.~Xiang.
\newblock Kaolin: A pytorch library for accelerating 3d deep learning research.
\newblock {\em arXiv preprint arXiv:1911.05063}, 2019.

\bibitem{joo2018total}
H.~Joo, T.~Simon, and Y.~Sheikh.
\newblock Total capture: A 3d deformation model for tracking faces, hands, and
  bodies.
\newblock In {\em Proceedings of the IEEE conference on computer vision and
  pattern recognition}, pages 8320--8329, 2018.

\bibitem{kanazawa2018end}
A.~Kanazawa, M.~J. Black, D.~W. Jacobs, and J.~Malik.
\newblock End-to-end recovery of human shape and pose.
\newblock In {\em Proceedings of the IEEE Conference on Computer Vision and
  Pattern Recognition}, pages 7122--7131, 2018.

\bibitem{kanazawa2019learning}
A.~Kanazawa, J.~Y. Zhang, P.~Felsen, and J.~Malik.
\newblock Learning 3d human dynamics from video.
\newblock In {\em Proceedings of the IEEE Conference on Computer Vision and
  Pattern Recognition}, pages 5614--5623, 2019.

\bibitem{kocabas2020vibe}
M.~Kocabas, N.~Athanasiou, and M.~J. Black.
\newblock Vibe: Video inference for human body pose and shape estimation.
\newblock In {\em Proceedings of the IEEE/CVF Conference on Computer Vision and
  Pattern Recognition}, pages 5253--5263, 2020.

\bibitem{kolotouros2019learning}
N.~Kolotouros, G.~Pavlakos, M.~J. Black, and K.~Daniilidis.
\newblock Learning to reconstruct 3d human pose and shape via model-fitting in
  the loop.
\newblock In {\em Proceedings of the IEEE International Conference on Computer
  Vision}, pages 2252--2261, 2019.

\bibitem{li2018implicit}
J.~Li, G.~Daviet, R.~Narain, F.~Bertails-Descoubes, M.~Overby, G.~E. Brown, and
  L.~Boissieux.
\newblock An implicit frictional contact solver for adaptive cloth simulation.
\newblock {\em ACM Transactions on Graphics (TOG)}, 37(4):1--15, 2018.

\bibitem{liu2019soft}
S.~Liu, T.~Li, W.~Chen, and H.~Li.
\newblock Soft rasterizer: A differentiable renderer for image-based 3d
  reasoning.
\newblock In {\em Proceedings of the IEEE International Conference on Computer
  Vision}, pages 7708--7717, 2019.

\bibitem{loper2015smpl}
M.~Loper, N.~Mahmood, J.~Romero, G.~Pons-Moll, and M.~J. Black.
\newblock Smpl: A skinned multi-person linear model.
\newblock {\em ACM transactions on graphics (TOG)}, 34(6):1--16, 2015.

\bibitem{luo2018orinet}
C.~Luo, X.~Chu, and A.~Yuille.
\newblock Orinet: A fully convolutional network for 3d human pose estimation.
\newblock In {\em BMVC}, 2018.

\bibitem{ma2020learning}
Q.~Ma, J.~Yang, A.~Ranjan, S.~Pujades, G.~Pons-Moll, S.~Tang, and M.~J. Black.
\newblock Learning to dress 3d people in generative clothing.
\newblock In {\em Proceedings of the IEEE/CVF Conference on Computer Vision and
  Pattern Recognition}, pages 6469--6478, 2020.

\bibitem{mehta2020xnect}
D.~Mehta, O.~Sotnychenko, F.~Mueller, W.~Xu, M.~Elgharib, P.~Fua, H.-P. Seidel,
  H.~Rhodin, G.~Pons-Moll, and C.~Theobalt.
\newblock Xnect: Real-time multi-person 3d motion capture with a single rgb
  camera.
\newblock {\em ACM Transactions on Graphics (TOG)}, 39(4):82--1, 2020.

\bibitem{mehta2017vnect}
D.~Mehta, S.~Sridhar, O.~Sotnychenko, H.~Rhodin, M.~Shafiei, H.-P. Seidel,
  W.~Xu, D.~Casas, and C.~Theobalt.
\newblock Vnect: Real-time 3d human pose estimation with a single rgb camera.
\newblock {\em ACM Transactions on Graphics (TOG)}, 36(4):1--14, 2017.

\bibitem{natsume2019siclope}
R.~Natsume, S.~Saito, Z.~Huang, W.~Chen, C.~Ma, H.~Li, and S.~Morishima.
\newblock Siclope: Silhouette-based clothed people.
\newblock In {\em Proceedings of the IEEE Conference on Computer Vision and
  Pattern Recognition}, pages 4480--4490, 2019.

\bibitem{omran2018neural}
M.~Omran, C.~Lassner, G.~Pons-Moll, P.~Gehler, and B.~Schiele.
\newblock Neural body fitting: Unifying deep learning and model based human
  pose and shape estimation.
\newblock In {\em 2018 international conference on 3D vision (3DV)}, pages
  484--494. IEEE, 2018.

\bibitem{paszke2019pytorch}
A.~Paszke, S.~Gross, F.~Massa, A.~Lerer, J.~Bradbury, G.~Chanan, T.~Killeen,
  Z.~Lin, N.~Gimelshein, L.~Antiga, et~al.
\newblock Pytorch: An imperative style, high-performance deep learning library.
\newblock In {\em Advances in Neural Information Processing Systems}, pages
  8024--8035, 2019.

\bibitem{pavlakos2019expressive}
G.~Pavlakos, V.~Choutas, N.~Ghorbani, T.~Bolkart, A.~A. Osman, D.~Tzionas, and
  M.~J. Black.
\newblock Expressive body capture: 3d hands, face, and body from a single
  image.
\newblock In {\em Proceedings of the IEEE Conference on Computer Vision and
  Pattern Recognition}, pages 10975--10985, 2019.

\bibitem{pavlakos2019texturepose}
G.~Pavlakos, N.~Kolotouros, and K.~Daniilidis.
\newblock Texturepose: Supervising human mesh estimation with texture
  consistency.
\newblock In {\em Proceedings of the IEEE International Conference on Computer
  Vision}, pages 803--812, 2019.

\bibitem{pavlakos2017coarse}
G.~Pavlakos, X.~Zhou, K.~G. Derpanis, and K.~Daniilidis.
\newblock Coarse-to-fine volumetric prediction for single-image 3d human pose.
\newblock In {\em Proceedings of the IEEE Conference on Computer Vision and
  Pattern Recognition}, pages 7025--7034, 2017.

\bibitem{pons2017clothcap}
G.~Pons-Moll, S.~Pujades, S.~Hu, and M.~J. Black.
\newblock Clothcap: Seamless 4d clothing capture and retargeting.
\newblock {\em ACM Transactions on Graphics (TOG)}, 36(4):1--15, 2017.

\bibitem{rhodin2016general}
H.~Rhodin, N.~Robertini, D.~Casas, C.~Richardt, H.-P. Seidel, and C.~Theobalt.
\newblock General automatic human shape and motion capture using volumetric
  contour cues.
\newblock In {\em European conference on computer vision}, pages 509--526.
  Springer, 2016.

\bibitem{robertini2018illumination}
N.~Robertini, F.~Bernard, W.~Xu, and C.~Theobalt.
\newblock Illumination-invariant robust multiview 3d human motion capture.
\newblock In {\em 2018 IEEE Winter Conference on Applications of Computer
  Vision (WACV)}, pages 1661--1670. IEEE, 2018.

\bibitem{robertini2016model}
N.~Robertini, D.~Casas, H.~Rhodin, H.-P. Seidel, and C.~Theobalt.
\newblock Model-based outdoor performance capture.
\newblock In {\em 2016 Fourth International Conference on 3D Vision (3DV)},
  pages 166--175. IEEE, 2016.

\bibitem{saito2019pifu}
S.~Saito, Z.~Huang, R.~Natsume, S.~Morishima, A.~Kanazawa, and H.~Li.
\newblock Pifu: Pixel-aligned implicit function for high-resolution clothed
  human digitization.
\newblock In {\em Proceedings of the IEEE International Conference on Computer
  Vision}, pages 2304--2314, 2019.

\bibitem{saito2020pifuhd}
S.~Saito, T.~Simon, J.~Saragih, and H.~Joo.
\newblock Pifuhd: Multi-level pixel-aligned implicit function for
  high-resolution 3d human digitization.
\newblock In {\em Proceedings of the IEEE/CVF Conference on Computer Vision and
  Pattern Recognition}, pages 84--93, 2020.

\bibitem{sigal2015perceptual}
L.~Sigal, M.~Mahler, S.~Diaz, K.~McIntosh, E.~Carter, T.~Richards, and
  J.~Hodgins.
\newblock A perceptual control space for garment simulation.
\newblock {\em ACM Transactions on Graphics (TOG)}, 34(4):1--10, 2015.

\bibitem{sun2018integral}
X.~Sun, B.~Xiao, F.~Wei, S.~Liang, and Y.~Wei.
\newblock Integral human pose regression.
\newblock In {\em Proceedings of the European Conference on Computer Vision
  (ECCV)}, pages 529--545, 2018.

\bibitem{tang2019neural}
S.~Tang, F.~Tan, K.~Cheng, Z.~Li, S.~Zhu, and P.~Tan.
\newblock A neural network for detailed human depth estimation from a single
  image.
\newblock In {\em Proceedings of the IEEE International Conference on Computer
  Vision}, pages 7750--7759, 2019.

\bibitem{varol2018bodynet}
G.~Varol, D.~Ceylan, B.~Russell, J.~Yang, E.~Yumer, I.~Laptev, and C.~Schmid.
\newblock Bodynet: Volumetric inference of 3d human body shapes.
\newblock In {\em Proceedings of the European Conference on Computer Vision
  (ECCV)}, pages 20--36, 2018.

\bibitem{volino2009simple}
P.~Volino, N.~Magnenat-Thalmann, and F.~Faure.
\newblock A simple approach to nonlinear tensile stiffness for accurate cloth
  simulation.
\newblock {\em ACM Transactions on Graphics (TOG)}, 28(4):1--16, 2009.

\bibitem{wei2016convolutional}
S.-E. Wei, V.~Ramakrishna, T.~Kanade, and Y.~Sheikh.
\newblock Convolutional pose machines.
\newblock In {\em Proceedings of the IEEE conference on Computer Vision and
  Pattern Recognition}, pages 4724--4732, 2016.

\bibitem{wu2013set}
C.~Wu, C.~Stoll, L.~Valgaerts, and C.~Theobalt.
\newblock On-set performance capture of multiple actors with a stereo camera.
\newblock {\em ACM Transactions on Graphics (TOG)}, 32(6):1--11, 2013.

\bibitem{xiang2019monocular}
D.~Xiang, H.~Joo, and Y.~Sheikh.
\newblock Monocular total capture: Posing face, body, and hands in the wild.
\newblock In {\em Proceedings of the IEEE Conference on Computer Vision and
  Pattern Recognition}, pages 10965--10974, 2019.

\bibitem{xu2018monoperfcap}
W.~Xu, A.~Chatterjee, M.~Zollh{\"o}fer, H.~Rhodin, D.~Mehta, H.-P. Seidel, and
  C.~Theobalt.
\newblock Monoperfcap: Human performance capture from monocular video.
\newblock {\em ACM Transactions on Graphics (ToG)}, 37(2):1--15, 2018.

\bibitem{xu2019denserac}
Y.~Xu, S.-C. Zhu, and T.~Tung.
\newblock Denserac: Joint 3d pose and shape estimation by dense
  render-and-compare.
\newblock In {\em Proceedings of the IEEE International Conference on Computer
  Vision}, pages 7760--7770, 2019.

\bibitem{yang2018analyzing}
J.~Yang, J.-S. Franco, F.~H{\'e}troy-Wheeler, and S.~Wuhrer.
\newblock Analyzing clothing layer deformation statistics of 3d human motions.
\newblock In {\em Proceedings of the European Conference on Computer Vision
  (ECCV)}, pages 237--253, 2018.

\bibitem{yu2018doublefusion}
T.~Yu, Z.~Zheng, K.~Guo, J.~Zhao, Q.~Dai, H.~Li, G.~Pons-Moll, and Y.~Liu.
\newblock Doublefusion: Real-time capture of human performances with inner body
  shapes from a single depth sensor.
\newblock In {\em Proceedings of the IEEE conference on computer vision and
  pattern recognition}, pages 7287--7296, 2018.

\bibitem{yu2019simulcap}
T.~Yu, Z.~Zheng, Y.~Zhong, J.~Zhao, Q.~Dai, G.~Pons-Moll, and Y.~Liu.
\newblock Simulcap: Single-view human performance capture with cloth
  simulation.
\newblock In {\em 2019 IEEE/CVF Conference on Computer Vision and Pattern
  Recognition (CVPR)}, pages 5499--5509. IEEE, 2019.

\bibitem{zhang2017detailed}
C.~Zhang, S.~Pujades, M.~J. Black, and G.~Pons-Moll.
\newblock Detailed, accurate, human shape estimation from clothed 3d scan
  sequences.
\newblock In {\em Proceedings of the IEEE Conference on Computer Vision and
  Pattern Recognition}, pages 4191--4200, 2017.

\bibitem{zheng2019deephuman}
Z.~Zheng, T.~Yu, Y.~Wei, Q.~Dai, and Y.~Liu.
\newblock Deephuman: 3d human reconstruction from a single image.
\newblock In {\em Proceedings of the IEEE International Conference on Computer
  Vision}, pages 7739--7749, 2019.

\bibitem{zhou2017towards}
X.~Zhou, Q.~Huang, X.~Sun, X.~Xue, and Y.~Wei.
\newblock Towards 3d human pose estimation in the wild: a weakly-supervised
  approach.
\newblock In {\em Proceedings of the IEEE International Conference on Computer
  Vision}, pages 398--407, 2017.

\bibitem{zhu2019detailed}
H.~Zhu, X.~Zuo, S.~Wang, X.~Cao, and R.~Yang.
\newblock Detailed human shape estimation from a single image by hierarchical
  mesh deformation.
\newblock In {\em Proceedings of the IEEE Conference on Computer Vision and
  Pattern Recognition}, pages 4491--4500, 2019.

\end{thebibliography}
}

\clearpage

\begin{strip}
\begin{center}
    {\Large \bf MonoClothCap: Towards Temporally Coherent Clothing Capture \\ from Monocular RGB Video (Supplementary Material) \par}
\end{center}
\end{strip}

\appendix
\section{Further Ablation Studies}

In this section, we conduct more ablation studies on various loss terms we use in the energy optimization for clothing capture and body shape estimation.

\subsection{Loss Terms for Clothing Capture}

We first study the loss terms used for clothing capture in Section 5.2 (Eq. 12). In the experiments below, we compare the results of the batch optimization stage with different loss terms, initialized from the same body capture and sequential tracking results.

\textbf{Clothing segmentation term (Eq. 13).} In order to study the effect of the clothing segmentation term, we run an ablative experiment where the weight for the segmentation term is set to $0$, while all other terms remain the same. To better visualize the effect, we render the output meshes in three colors: grey for skin, yellow for upper clothing and green for lower clothing. We consider a vertex $j$ as a skin vertex if the length of the clothing offset for this vertex is below a certain threshold $\varepsilon$, or
$$ \Vert D_j \Vert < \varepsilon, $$
where $D_j$ is defined in Eq. 4 in the main paper. We consider a vertex as belonging to the upper clothing if
$$ \Vert D_j^u \Vert \ge \Vert D_j^l \Vert \quad \text{and} \quad \Vert D_j \Vert \ge \varepsilon, $$
or, similarly, as belong to the lower clothing if
$$ \Vert D_j^l \Vert > \Vert D_j^u \Vert \quad \text{and} \quad \Vert D_j \Vert \ge \varepsilon. $$

\begin{figure*}[h]
  \centering
  \includegraphics[width=\linewidth]{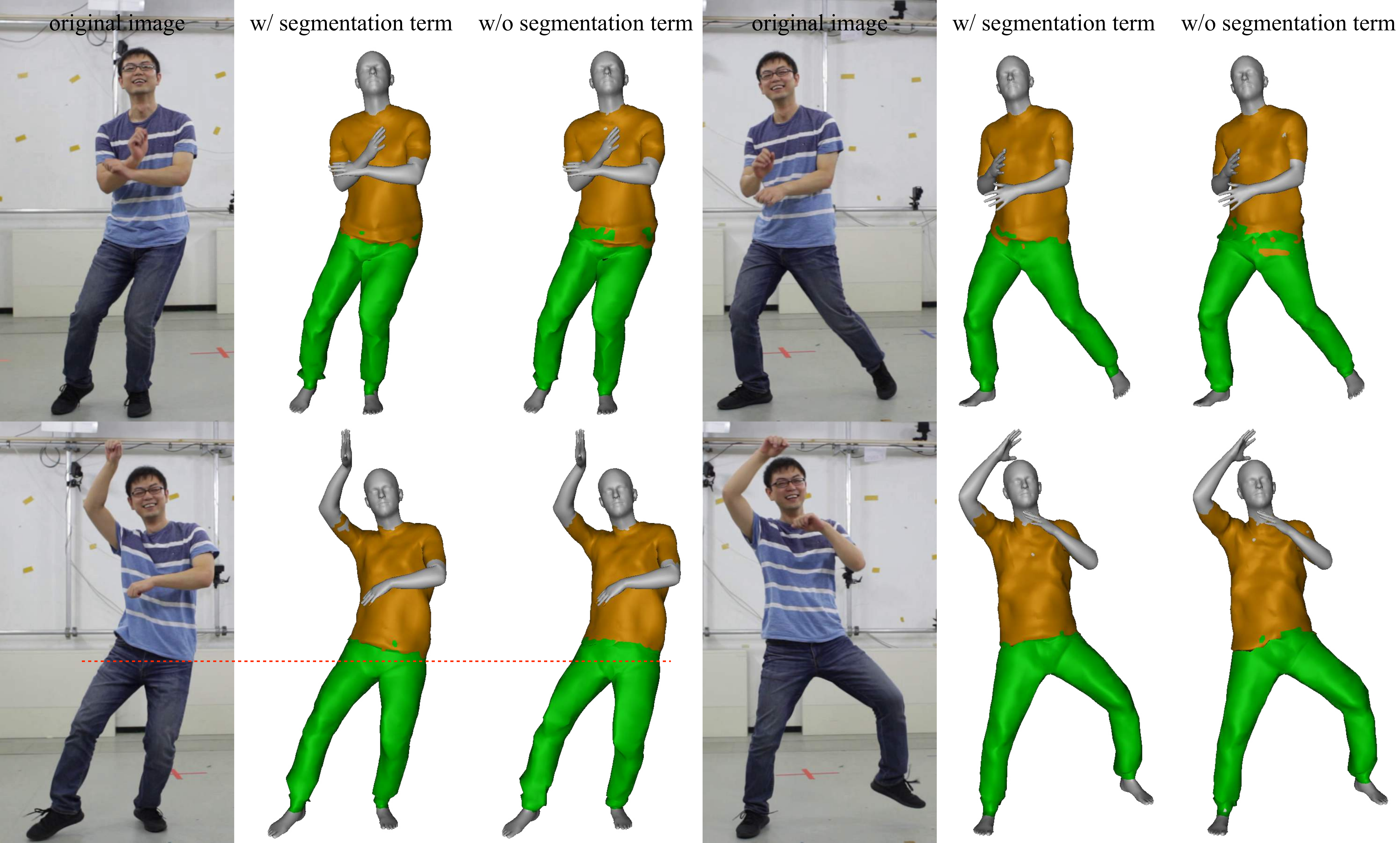}
  \caption{Comparison between results with and without clothing segmentation loss. The vertices for skin, upper clothing and lower clothing are rendered in grey, yellow and green respectively. A horizontal red dashed line is drawn in the bottom left example to help visually check the location of the boundary between upper and lower clothing.}
  \label{fig:supp-ablation-seg}
\end{figure*}

The result of this experiment is shown in Fig.~\ref{fig:supp-ablation-seg}. In each frame, we observe that the boundary between the upper and lower clothing is \textit{more consistent} with the original image in the result with segmentation term than the result without segmentation term. Our method adopts a combination of upper clothing and lower clothing models, which might both have non-zero offsets around the body waist. It is important for our method to produce both offsets with correct relative length to realistically reconstruct the spatial arrangement of the T-shirt and trousers in the original images. This result proves the effectiveness and necessity of the clothing segmentation term.

\textbf{Photometric tracking term (Eq. 14).} Similarly, we run an ablative experiment where the weight for the photometric tracking term is set to $0$ and other terms remain the same. To visualize its effect, we render the output tracked mesh with the final texture extracted in the sequential tracking stage (see Section 5.2 of the main paper for detail), and compare the results with and without the photometric tracking term with the original images. 

\begin{figure*}[h]
  \centering
  \includegraphics[width=0.95\linewidth]{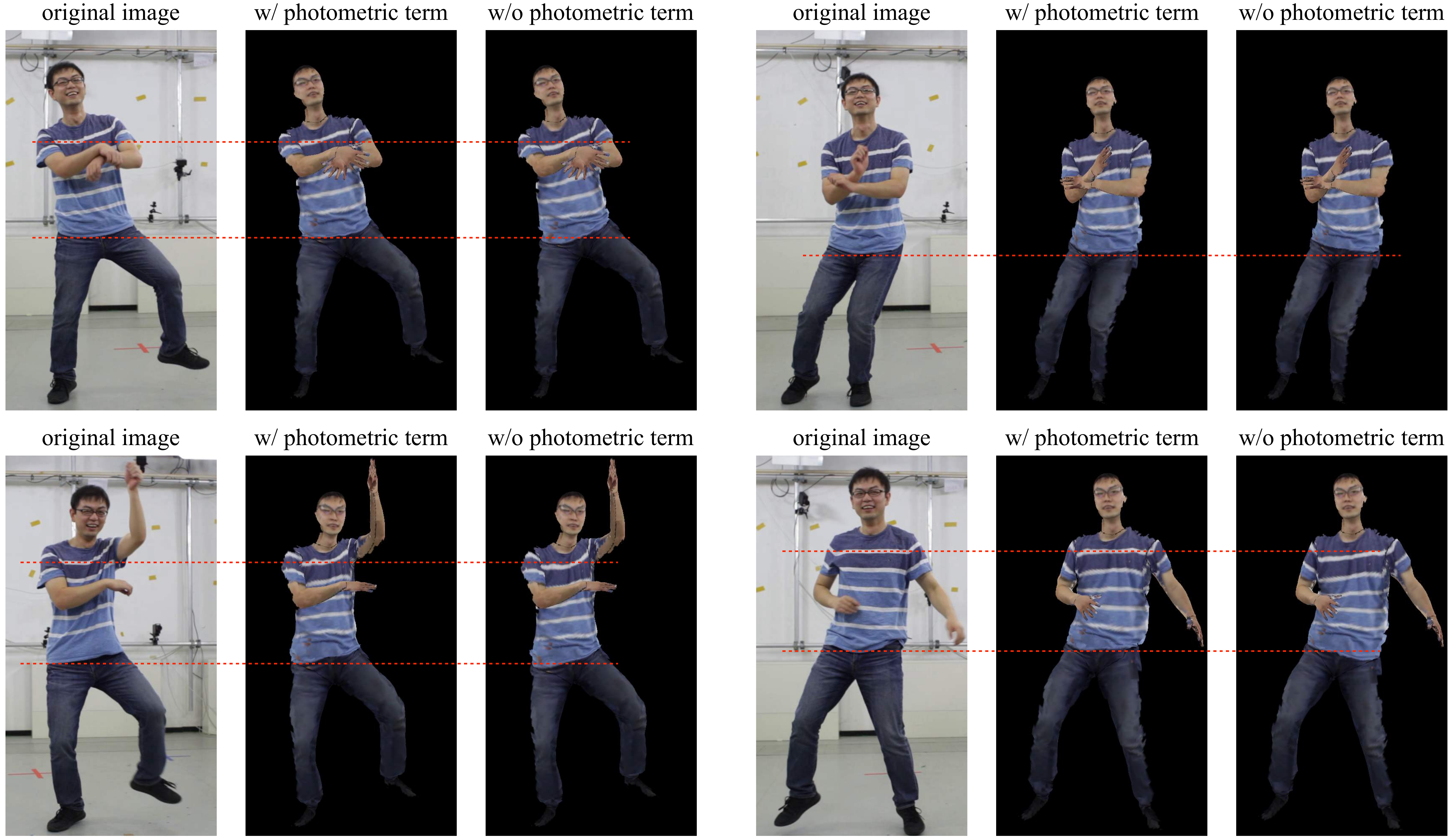}
  \caption{Comparison between results with and without photometric tracking loss. Horizontal dashed lines are drawn in red to help visually compare the location of rendered texture pattern.}
  \label{fig:supp-ablation-photometric}
\end{figure*}

The result of this experiment is shown in Fig.~\ref{fig:supp-ablation-photometric}. Notice that the same final texture image is used to render all the results. In order to assist visual comparison of the rendered pattern, we draw several auxiliary horizontal dashed lines in red. We can observe that the results with photometric tracking term is more consistent with the original image than the result without photometric tracking term, in terms of the location of the white strip on the T-shirt and the boundary between the T-shirt and trousers. This demonstrates that our photometric tracking loss can help to obtain better temporal correspondence across different frames in the video.

\textbf{Silhouette matching term.} We now compare the results with and without the silhouette matching term. We render both results and align them with the original images to visualize how well the silhouette matches.

\begin{figure*}[h]
  \centering
  \includegraphics[width=0.95\linewidth]{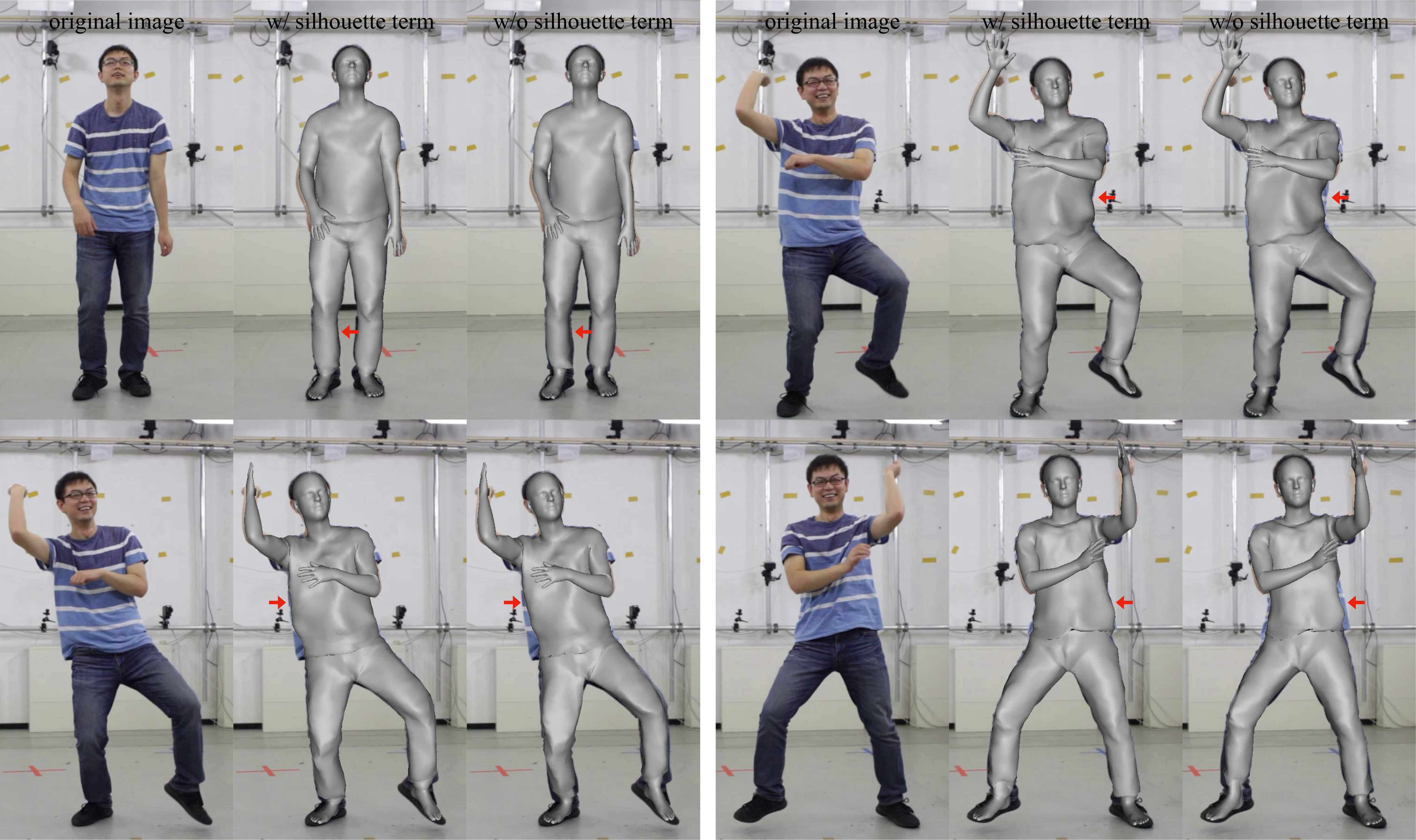}
  \caption{Comparison between results with and without silhouette matching loss. In each example, we show the original image, the result with and without silhouette matching loss from left to right.}
  \label{fig:supp-ablation-sil}
\end{figure*}

The result of this experiment is shown in Fig.~\ref{fig:supp-ablation-sil}. We observe that the result with silhouette matching term achieves a better alignment of silhouette with the original image. This suggests that the silhouette matching term can help to reconstruct the accurate shape of the clothing in the video.

\subsection{Losses Terms for Body Shape Estimation}

Although body pose and shape estimation is not a focus of this paper, we conduct ablative studies on the loss terms used in body shape estimation in Section 5.1. (Eq. 9). In each of the experiment in this section, the weight for the loss term under study is set to $0$, and all other terms stay the same as the full results. We render the estimated body shapes and compare them with the full results.

\begin{figure*}[h]
  \centering
  \includegraphics[width=0.95\linewidth]{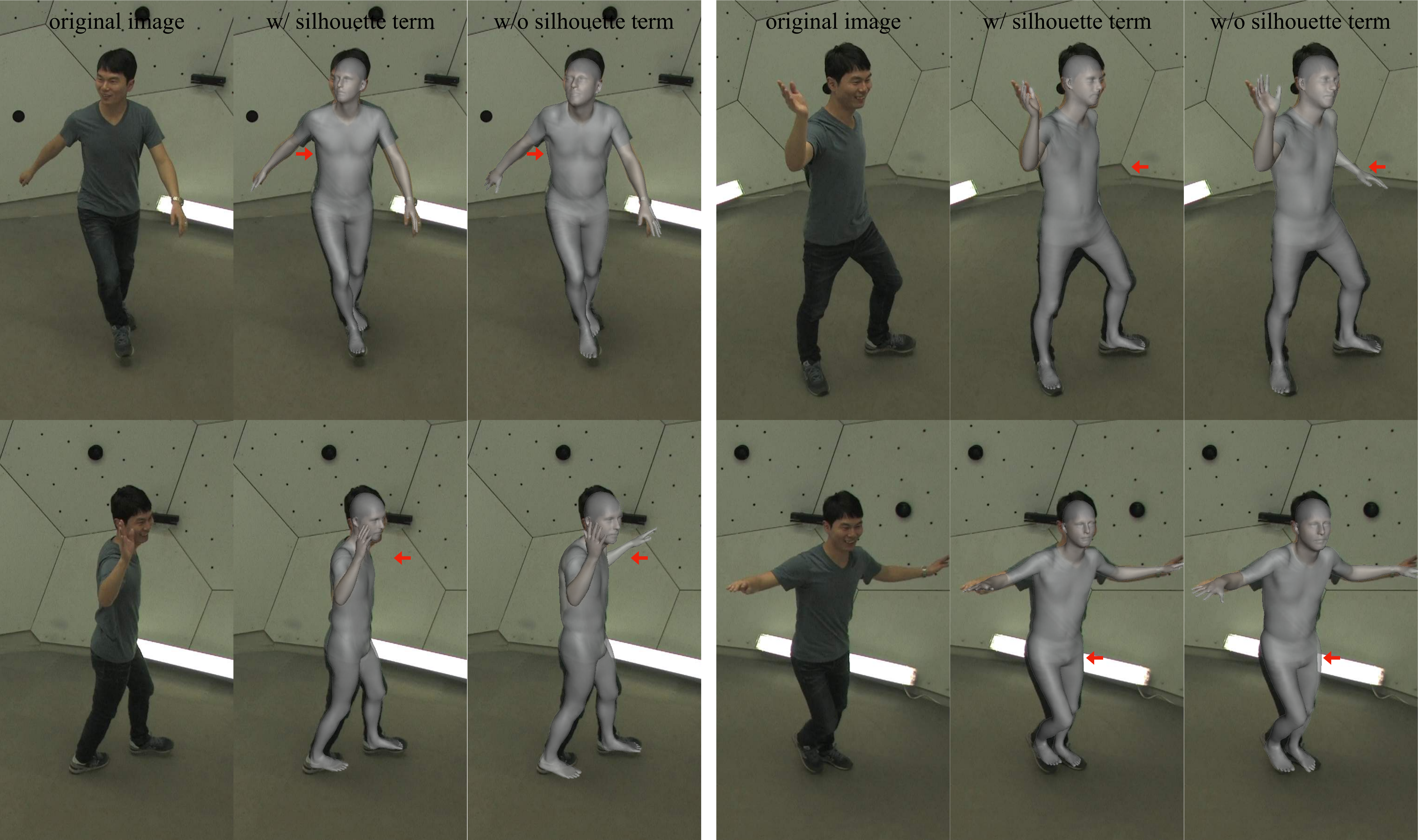}
  \caption{Comparison between results with and without silhouette loss. In each example, we show the original image, the result with and without silhouette loss from left to right.}
  \label{fig:supp-ablation-body-sil}
\end{figure*}

\textbf{Silhouette term.} The result of this experiment is shown in Fig.~\ref{fig:supp-ablation-body-sil}. We can observe in the result that silhouette provides critical information for the estimation of body shape and pose in the following two ways. First, the projection of human body should always lie in the interior of the overall silhouette in the image, which includes the region of body and clothes. Second, in the top-right and bottom-left examples, an arm of the subject is occluded by the torso. There is no available information to reason about the location of the arm from the 2D keypoints or DensePose results. In this situation, only the silhouette can constrain the position of the arm to be behind the torso in the camera view. This proves the importance of the silhouette term for accurate estimation of human body and shape.


\begin{figure*}[h]
  \centering
  \includegraphics[width=0.95\linewidth]{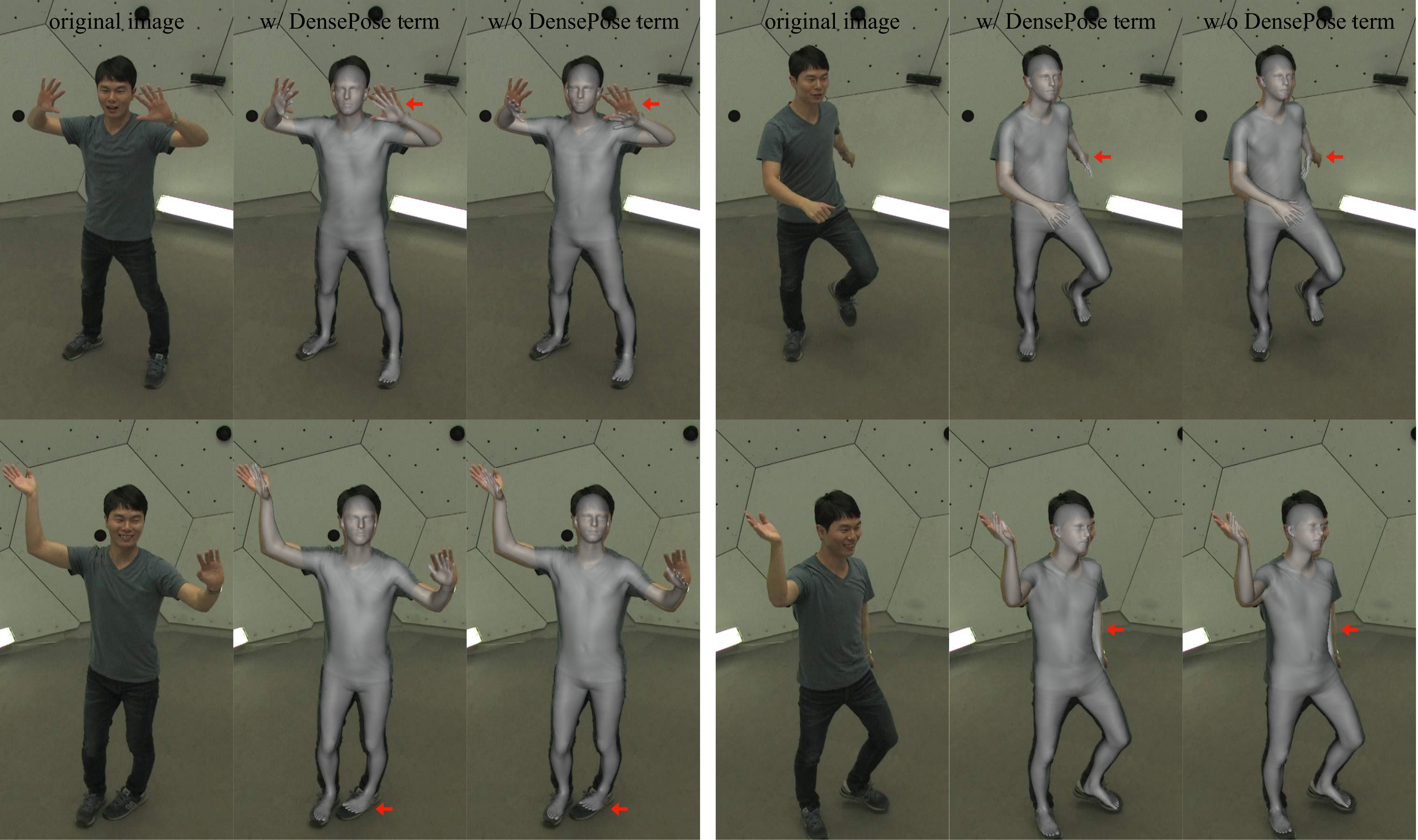}
  \caption{Comparison between results with and without DensePose loss. In each example, we show the original image, the result with and without DensePose loss from left to right.}
  \label{fig:supp-ablation-body-dp}
\end{figure*}

\textbf{DensePose term.} The result of this experiment is shown in Fig.~\ref{fig:supp-ablation-body-dp}. The use of DensePose together with SMPL model for accurate body estimation was first proposed in \cite{guler2019holopose}. In our work, we find that the DensePose term helps to estimate the hand orientation more accurately, as fingers are usually not included in the hierarchy of 2D body pose output.

\begin{figure*}[h]
  \centering
  \includegraphics[width=0.95\linewidth]{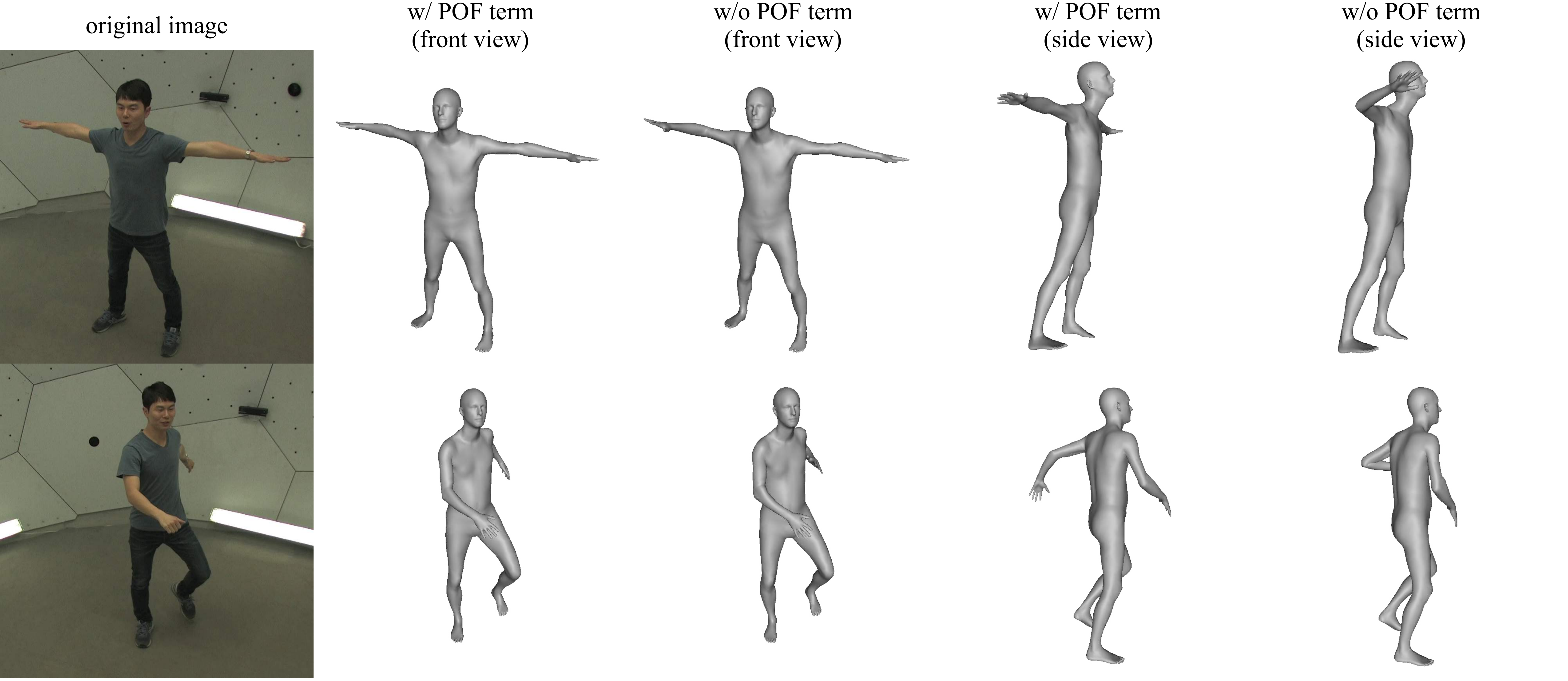}
  \caption{Comparison between results with and without POF loss. In each example, we show the original image, the result with and without POF loss from both the front view and the side view.}
  \label{fig:supp-ablation-body-POF}
\end{figure*}

\textbf{POF term.} The result of this experiment is shown in Fig.~\ref{fig:supp-ablation-body-POF}. The use of POF together with deformable human body model was first proposed in \cite{xiang2019monocular}. We find that the POF term can help to eliminate the ambiguity of 3D body pose given only 2D keypoints in the front view, and therefore help to estimate more accurate body pose in 3D.

\section{Quantitative Comparison with Monocular 3D Pose Estimation Methods}

In the first stage of our pipeline, we use a standard model-fitting method to estimate 3D body pose from the video. Although we do not claim any contribution or novelty in this aspect, we still provide a quantitative comparison with recent state-of-the-art approaches that estimate 3D body pose with SMPL model from a monocular view. In particular, we evaluate all methods on the \textit{Pablo} sequence using the same protocol as Section 6.1 in the main paper. The evaluation results are shown in Table \ref{tab:3dpose}. As a part of our pipeline, our estimation of 3D body pose is highly accurate even when compared with recent state-of-the-art approaches that focus on 3D body pose only. This lays a solid foundation for the following clothing capture stages.

\begin{table}[t]
\centering
\begin{tabular}{c  c  c  c}
    \toprule
    Ours & Temporal HMR \cite{kanazawa2019learning} & SPIN \cite{kolotouros2019learning} & VIBE \cite{kocabas2020vibe} \\
    \textbf{77.3} & 94.7 & 89.5 & 87.2 \\
    \bottomrule
\end{tabular}
\caption{Quantitative comparison with recent SMPL-based 3D body pose estimation approaches on the \textit{Pablo} sequence. All numbers are in mm.}
\label{tab:3dpose}
\end{table}

\section{Runtime Analysis}

In this section, we present the runtime information of our approach. Our method runs on a Linux server with 40 CPU cores and 4 GTX TITAN X GPUs. Our approach requires the memory of 4 GPUs in order to run the batch optimization on a video of around 250 frames together. For optimization, we use the L-BFGS solver implemented in PyTorch \cite{paszke2019pytorch}. We measure the average time consumed for each frame in every stage, and the results are shown in Table \ref{tab:runtime}.

\begin{table}[t]
\centering
\begin{tabular}{c c}
    \toprule
    Stage & Runtime (s) \\
    Body Estimation (Sec. 5.1) & 6 \\
    Sequential Tracking (Sec. 5.2) & 62 \\
    Batch Optimization (Sec. 5.2) & 27 \\
    Wrinkle Extraction (Sec. 5.3) & 232 \\
    Total & 327 \\
    \bottomrule
\end{tabular}
\caption{Average per-frame runtime of each stage in our pipeline. The numbers are in seconds.}
\label{tab:runtime}
\end{table}

\section{Complete Quantitative Evaluation Results}

In this section, we present the figures for complete per-frame results of the quantitative experiments conducted in Section 6 of the main paper.

\subsection{Evaluation on MonoPerfCap Dataset}

\begin{figure*}[ht]
  \centering
  \includegraphics[width=0.6\linewidth]{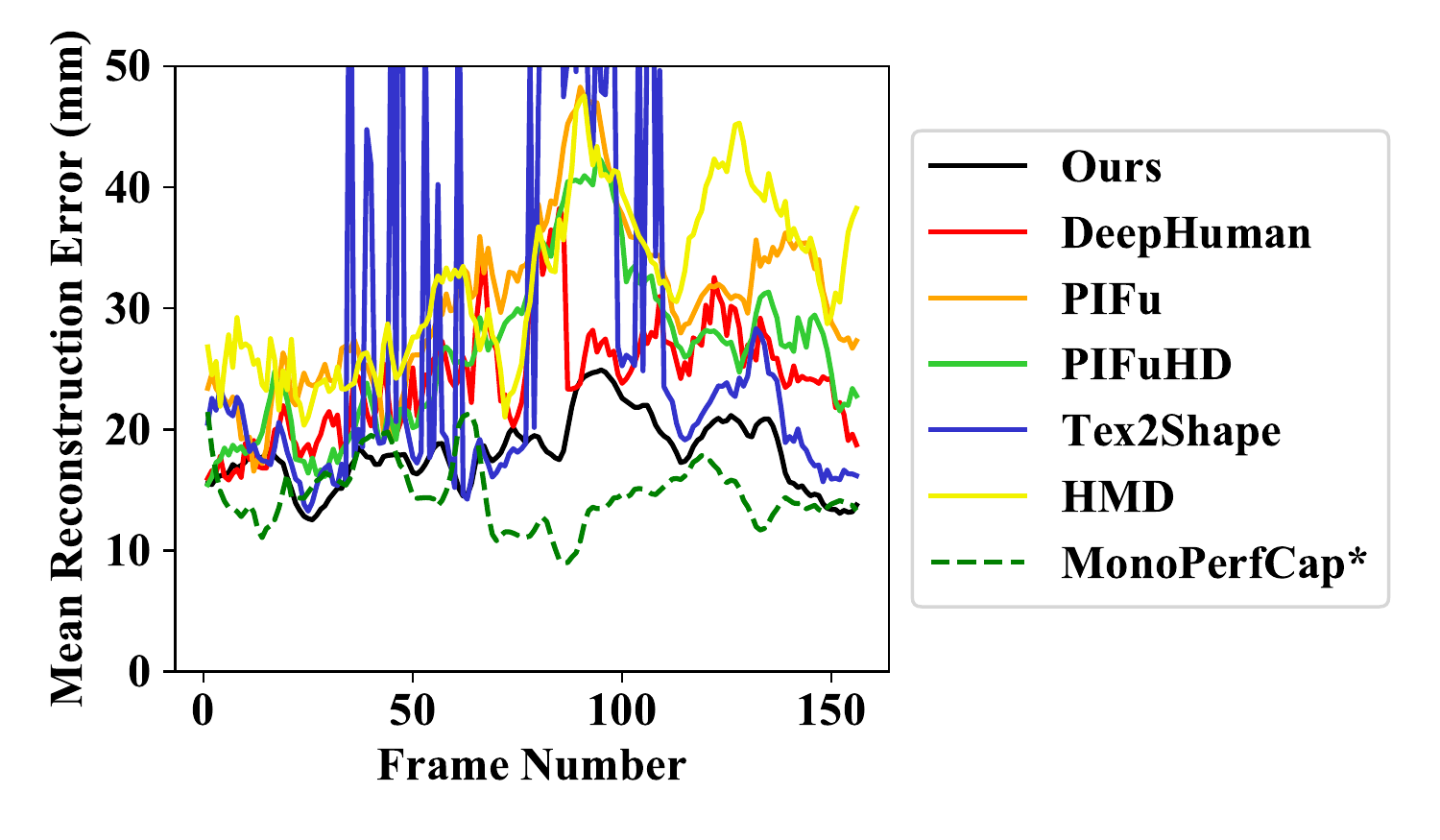}
  \caption{Per-frame results of the quantitative comparison with previous work on \textit{Pablo} sequence using mean point-to-surface error. Notice that the method annotated with `*' uses a pre-scanned personalized template that provides strong shape prior.}
  \label{fig:supp-pablo-surface}
\end{figure*}

\textbf{Evaluation of Clothing Surface Reconstruction.} The complete per-frame results corresponding to the surface error in Table 1 in the main paper are shown in Fig.~\ref{fig:supp-pablo-surface}.

\begin{figure*}[ht]
  \centering
  \includegraphics[width=0.6\linewidth]{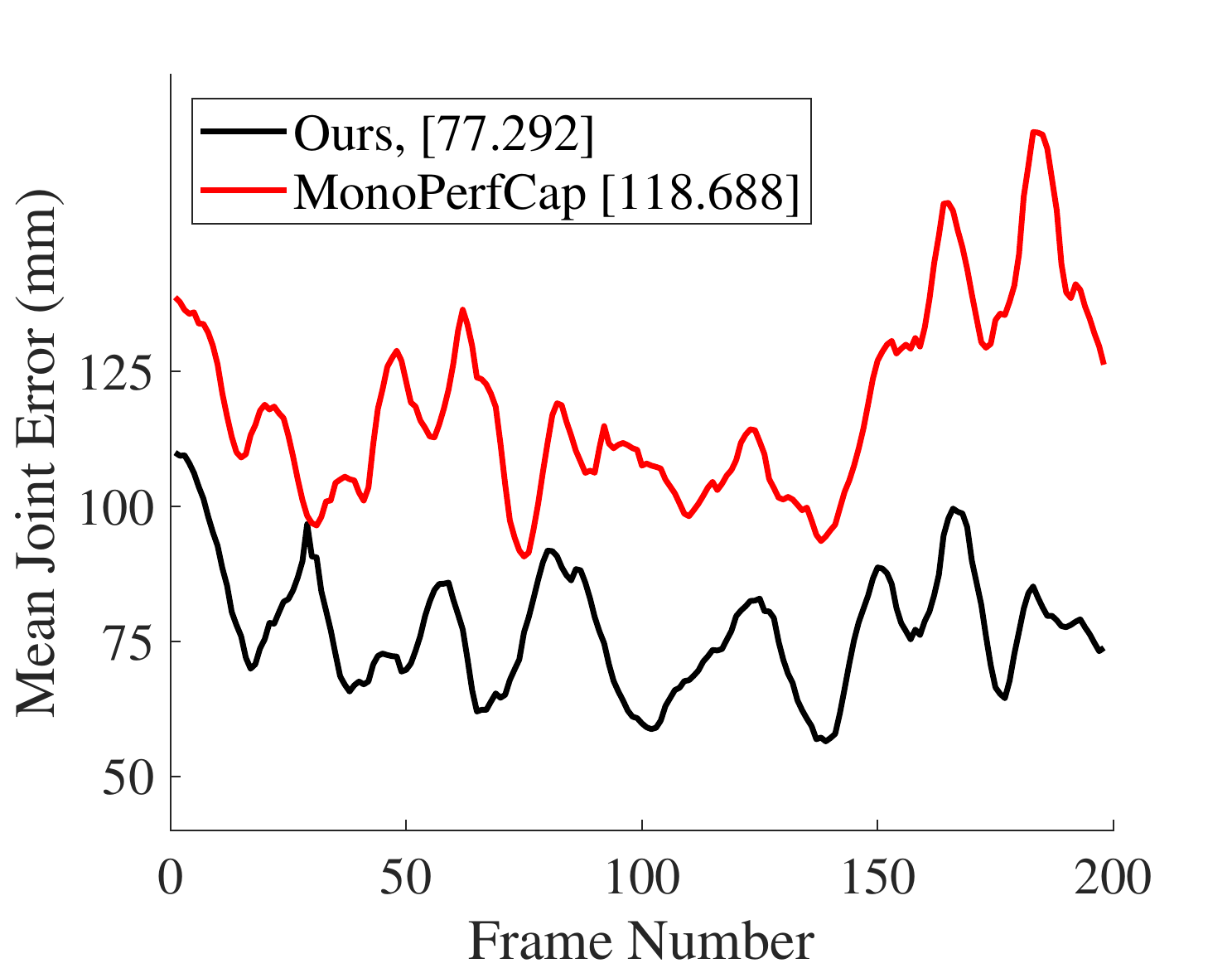}
  \caption{Per-frame results of the quantitative comparison with previous work on \textit{Pablo} sequence using mean joint error.}
  \label{fig:supp-pablo-joints}
\end{figure*}

\textbf{Evaluation of 3D Pose Estimation.} The complete per-frame results corresponding to the joint error in Table 1 in the main paper are shown in Fig.~\ref{fig:supp-pablo-joints}.

\subsection{Evaluation on BUFF Dataset}

\begin{figure*}[ht]
  \centering
  \includegraphics[width=\linewidth]{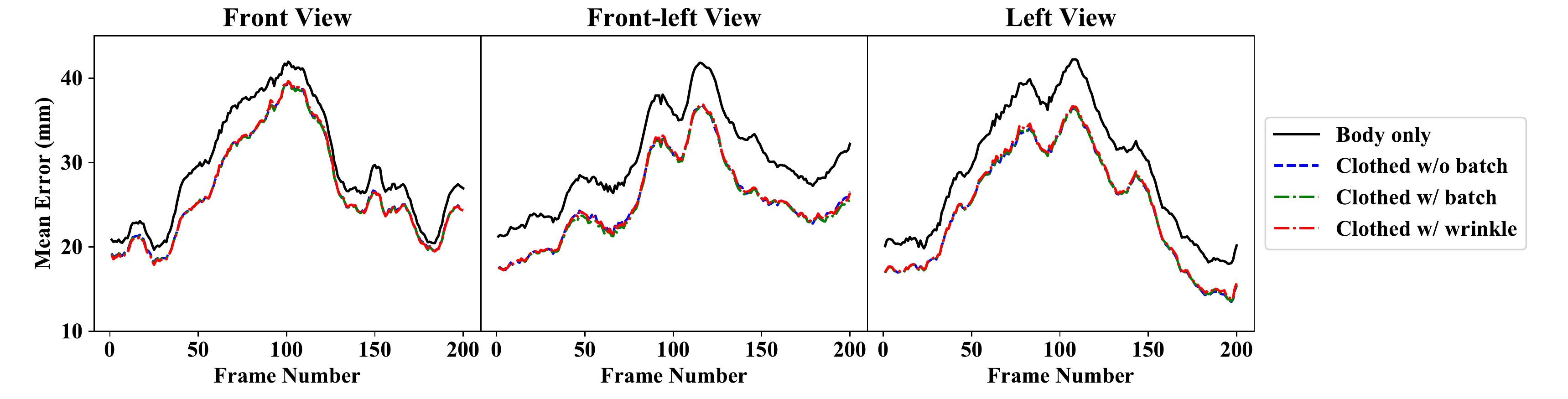}
  \caption{Per-frame results of the quantitative ablation study for different stages of our method on rendered BUFF dataset using mean point-to-surface error.}
  \label{fig:supp-buff}
\end{figure*}

The complete per-frame results corresponding to Table 2 in the main paper are shown in Fig.~\ref{fig:supp-buff}.

\end{document}